\documentclass{article}

\usepackage{microtype}
\usepackage{graphicx}
\usepackage{subcaption}
\usepackage{booktabs} 

\usepackage{hyperref}
\usepackage{url}

\usepackage[preprint]{icml2026}

\usepackage{amsmath}
\usepackage{amssymb}
\usepackage{mathtools}
\usepackage{amsthm}

\usepackage[capitalize,noabbrev]{cleveref}

\usepackage{multirow}

\theoremstyle{plain}
\newtheorem{theorem}{Theorem}[section]

\theoremstyle{definition}
\newtheorem{definition}[theorem]{Definition}

\theoremstyle{remark}

\usepackage[textsize=tiny]{todonotes}

\begin{document}

\twocolumn[
  \icmltitle{Dimensional Collapse in Transformer Attention Outputs: A Challenge for Sparse Dictionary Learning}



  \icmlsetsymbol{equal}{*}

  \begin{icmlauthorlist}
    \icmlauthor{Junxuan Wang}{yyy,sch}
    \icmlauthor{Xuyang Ge}{yyy,sch}
    \icmlauthor{Wentao Shu}{yyy,sch}
    \icmlauthor{Zhengfu He}{yyy,sch}
    \icmlauthor{Xipeng Qiu}{yyy,sch}
  \end{icmlauthorlist}

  \icmlaffiliation{yyy}{Shanghai Innovation Institute, Shanghai, China}
  \icmlaffiliation{sch}{OpenMOSS Team, School of Computer Science, Fudan University, Shanghai, China}

  \icmlcorrespondingauthor{Xipeng Qiu}{xpqiu@fudan.edu.cn}

  \icmlkeywords{Sparse Autoencoders, Understanding high-level properties of models, Transformer}

  \vskip 0.3in
]



\printAffiliationsAndNotice{}  

\begin{figure*}[h]
    \centering
    \includegraphics[width=0.95\textwidth]{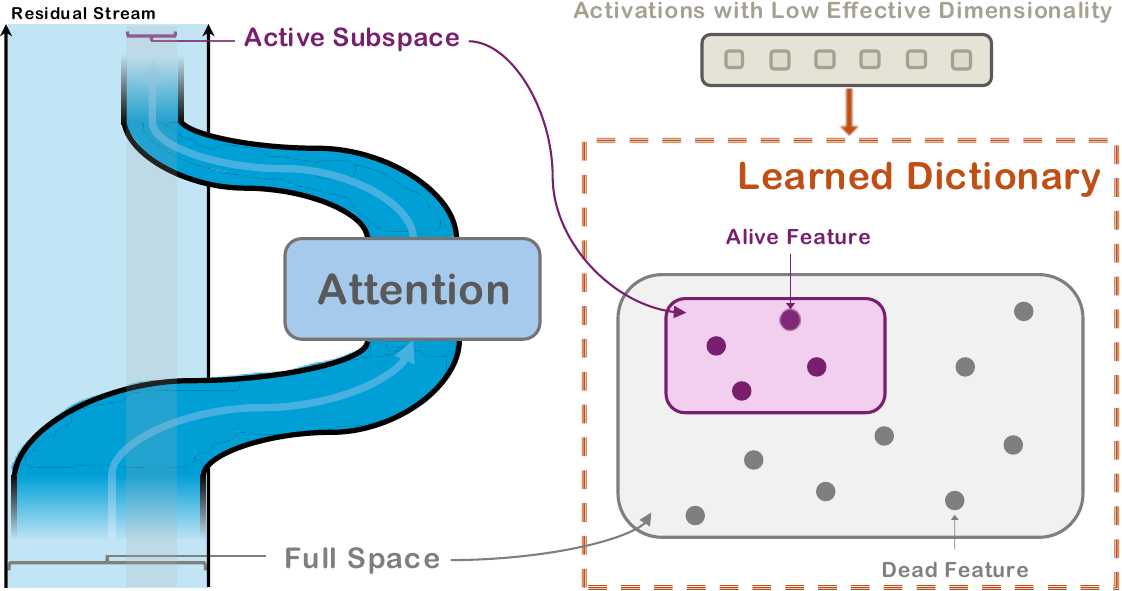}
    \caption{(left) Attention outputs exhibit pronounced low-rank structure compared to residual streams and MLP outputs, indicating that the attention layer writes into a subspace of the residual stream. (right) Low effective dimensionality of activations is a root cause of dead features in sparse dictionary learning methods. Setting feature directions in the active subspace mitigates this issue.}
    \label{fig:head_pic}
\end{figure*}

\begin{abstract}
Transformer architectures, and their attention mechanisms in particular, form the foundation of modern large language models. While transformer models are widely believed to operate in high-dimensional hidden spaces, we show that attention outputs are in fact confined to a surprisingly low-dimensional subspace, with an effective dimensionality of only about $60\%$ of the full space.
In contrast, MLP outputs and residual streams remain much closer to full-rank, exhibiting effective ranks around $90\%$.
This striking dimensional discrepancy is consistently observed across diverse model families and datasets, and is strongly shaped by the attention output projection matrix. Critically, we find this low-rank structure as a key factor of the prevalent dead feature problem in sparse dictionary learning, where it creates a mismatch between randomly initialized features and the intrinsic geometry of the activation space. Building on this insight, we propose a subspace-constrained training method for sparse autoencoders (SAEs), initializing feature directions into the active subspace of activations. Our approach reduces dead features from 87\% to below 1\% in Attention Output SAEs with 1M features, and can further extend to other sparse dictionary learning methods. Our findings provide both new insights into the geometry of attention and practical tools for improving sparse dictionary learning in large language models. 
\end{abstract}

\section{Introduction}

Over the past years, mechanistic interpretability has shifted from a collection of proof-of-concept tools~\citep{olsson2022context, wang2022ioi, meng2023locatingeditingfactualassociations, gould2023successorheads} toward a fast-growing, scale-driven field~\citep{ameisen2025circuit, lindsey2025biology}. This transformation is driven by a wave of sparse dictionary learning methods--such as sparse autoencoders (SAEs) and their variants~\citep{cunningham2023sae, bricken2023monosemanticity, lindsey2024crosscoder}, transcoders~\citep{dunefsky2024transcoder, ge2024hierattr}, and low-rank sparse attention~\citep{he2025lorsa}--that once targeted small models but are now being pushed to larger architectures and wider model families~\citep{templeton2024scaling, gao2024oaisae,hazra2025deepseekr1sae}. As these approaches scale in performance and model coverage, they provide increasingly complete and fine-grained explanations of neural network behavior~\citep{lindsey2024saescalinglaw, gao2024oaisae}.

However, scaling these approaches presents substantial practical challenges~\citep{templeton2024scaling, gao2024oaisae, mudide2025switchsae}. As models and feature dictionaries grow, the parameter count increases rapidly, leading to significant computational overhead. Moreover, a large fraction of learned features remain inactive, resulting in considerable waste in both computation and memory~\citep{templeton2024scaling, kissane2024attentionsaelesswrong}. In this work, we identify \textbf{the low-rank structure of the activations as a primary driver of dead features}~(Section~\ref{sec:leveraging_subspace_improve_sae:low_rank_dead_cooccur}).

Through singular value decomposition and effective dimensionality analyses~\citep{roy2007erank, staats2025smallsingularvaluesmatter}, we show for the first time that \textbf{the outputs of multi-head self-attention in transformer-based language models exhibit a remarkably low-rank structure}~(Section~\ref{sec:activation_spectra}). Compared to multilayer perceptron~(MLP) outputs and residual streams, attention outputs consistently concentrate in a low-dimensional subspace. We show that this behavior is robust across layers, datasets, and model families, including GPT-2~\citep{radford2019gpt2}, Llama~3.1~\citep{dubey2024llama3}, Gemma~2~\citep{rivire2024gemma2}, and Qwen~3~\citep{yang2025qwen3}. This universality aligns with prior observations of shared structures across models~\citep{olah2020zoom, chughtai2023toymodeluniv, gurnee2024universalneurons, wang2025univ}, while revealing a distinct form of representation collapse~\citep{hua2021featuredecorrelation, jing2022understanddc} at the level of attention outputs. We trace the origin of this low-rank structure to the anisotropy of the output projection matrix $W^O$, which further compresses multi-head activations into a lower-dimensional subspace.

In Section~\ref{sec:leveraging_subspace_improve_sae}, we investigate how the low-rank structure of attention outputs interacts with SAE training.  
By evaluating the full suite of open-source SAEs from \textit{LlamaScope}~\citep{he2024llamascope}, we show that low effective dimensionality strongly correlates with the number of dead features, suggesting a mismatch between random initialization and the low-dimensional geometry of the activations.
Drawing inspiration from \citet{phan2025principalcomponentsinit}'s principal component initialization for the first network layer, we propose \textit{Active Subspace Initialization}, which aligns SAE features with the active subspace of activations, \textbf{substantially reducing dead features while improving reconstruction}. 
Following~\citet{lindsey2024saescalinglaw} and~\citet{gao2024oaisae}, we conduct scaling experiments, which further reveal that ASI achieves superior reconstruction across feature counts, and when combined with SparseAdam\footnote{See the \href{https://docs.pytorch.org/docs/stable/generated/torch.optim.SparseAdam.html}{SparseAdam documentation} for details.}, it achieves the best reconstruction in large scale and reduces dead features from 87\% to below 1\% in Attention Output SAEs with 1M features trained on Llama-3.1-8B~\citep{dubey2024llama3}.

Furthermore, we show that \textbf{Active Subspace Init can generalize to sparse replacement models}~\citep{he2025lorsa, dunefsky2024transcoder, ameisen2025circuit}~(Section~\ref{sec:leveraging_subspace_improve_sae:general_to_lorsa}). When applied to other sparse dictionary learning methods, our initialization procedure systematically reduces the prevalence of dead parameters across architectures.

\section{Related Work}
\label{sec:related_work}

\subsection{Representation Collapse in Neural Representations and Low-Rankness in Attention Mechanism}
\label{sec:related_work:low_rank_in_attention}

A long line of research has shown that neural representations frequently concentrate in low-dimensional subspaces, forming representation collapse~\citep{hua2021featuredecorrelation, tian2021understandssld, jing2022understanddc}. These works generally focus on visual models trained using self-supervised methods.

Within attention mechanisms, prior work has investigated various notions of ``low-rankness'': low-rank approximation of attention patterns \citep{wang2020linformer, tay2020synthesizer, raganato2020fixedencoderattention}, low-rank parameterization for model compression \citep{noach2020matdecomposition,hu2022lora}, and the inherent low-rank bottleneck in single-head outputs \citep{bhojanapalli2020lowrankbottleneck}. 

Different from these prior lines of work, we demonstrate that the multi-head self-attention outputs exhibit a low-rank structure, revealing a distinct and under-explored phenomenon.

\subsection{Superposition Hypothesis and Sparse Dictionary Learning Methods}
\label{sec:related_work:sparse_dict_learning_methods}

The superposition hypothesis posits that neurons encode multiple non-orthogonal underlying features~\citep{arora2018linearword, olah2020zoom, elhage2022superposition, park2024llmgeometry}. Motivated by this view, a variety of sparse dictionary learning methods have been developed for interpretability, including sparse autoencoders and their variants~\citep{cunningham2023sae, bricken2023monosemanticity, lindsey2024crosscoder}, transcoders~\citep{dunefsky2024transcoder, ge2024hierattr}, and low-rank sparse attention~\citep{he2025lorsa}. These approaches decompose activations into sparse combinations of learned features while differing in their mechanisms for predicting or approximating feature activations. Their successful application across a wide range of model scales~\citep{templeton2024scaling, lieberum2024gemmascope, he2024llamascope}, architectures~\citep{wang2025univ}, and modalities~\citep{abdulaal2024xraysae} highlights their practical effectiveness for interpretability; however, they do not constitute direct hypothesis tests of the superposition hypothesis, which remain active topics of debate~\citep{sharkey2025open}.

\subsection{Dead Features in Sparse Dictionary Learning Methods}
\label{sec:related_work:root_origin_of_dead}

A persistent challenge in sparse dictionary learning methods is the emergence of \emph{dead features}\footnote{Following \citet{bricken2023monosemanticity}, we define a feature as dead if it never activates over 10 million tokens in this paper.}~\citep{templeton2024scaling, kissane2024attentionsaelesswrong}, which are also referred to as \emph{dead units} in sparse replacement models~\citep{dunefsky2024transcoder, ge2024hierattr, he2025lorsa}.
These features contribute nothing to reconstruction quality, wasting parameters and computation. Existing approaches to mitigate this issue rely on auxiliary loss terms~\citep{gao2024oaisae, conerly2025dltechnique} or resampling strategies~\citep{bricken2023monosemanticity} to encourage feature usage.

\subsection{PCA-Inspired Network Initialization}
\label{sec:related_work:pca_init}

A common practice applies PCA to input data for dimensionality reduction before network training~\citep{hastie2009statisticallearning, montavon2012tricksoftrade, jolliffe1986pca, bishop2007prml}. Recently, \citet{phan2025principalcomponentsinit} proposed \emph{PCsInit}, which initializes the first layer weights of networks with top principal components of data—embedding the PCA transform directly into the network. This provides the model with a superior parameter set~\citep{gu2025advantageousparameterexpansiontraining}, boosting performance by construction.

\begin{figure*}[t]
    \centering
    \includegraphics[width=\textwidth]{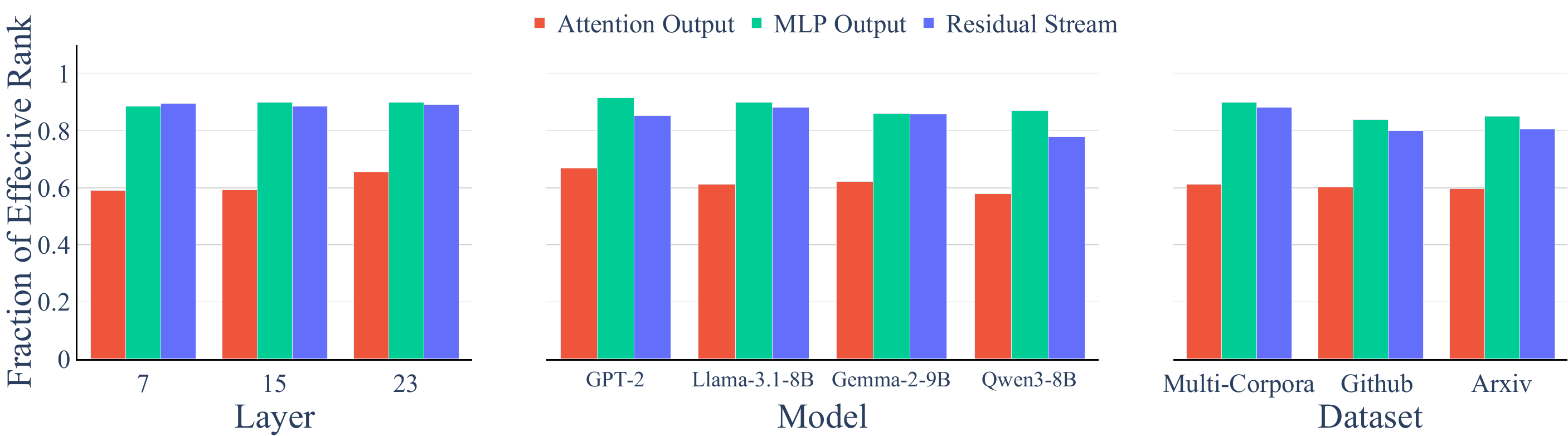}
    \caption{
    Across layers, model families and datasets, \textcolor[HTML]{EF553B}{attention outputs} exhibit dramatically lower effective rank than \textcolor[HTML]{636EFA}{residual streams} and \textcolor[HTML]{00CC96}{MLP outputs}, indicating that the attention layer writing into a low dimensional subspace of residual stream is a universal phenomenon. Details in Section~\ref{sec:activation_spectra:svd}.
    (left) Evaluation of Llama-3.1-8B on SlimPajama~\citep{cerebras2023slimpajama} dataset.
    (mid) Middle-layer analysis across model families on SlimPajama dataset.
    (right) Middle-layer analysis of Llama-3.1-8B across datasets.
    }
    \label{fig:effective_rank}
\end{figure*}

\section{Preliminaries}
\label{sec:preliminary}

\subsection{Multi-Head Self-Attention and Notations}
\label{sec:preliminary:mhsa}

We consider a Transformer block with multi-head self-attention (MHSA)~\citep{vaswani2017transformer}. Given input activations \(X \in \mathbb{R}^{n \times d}\), where \(n\) is the token count and \(d\) is the model hidden size, each attention head \(i\) computes:
\begin{equation*}
\small
\begin{aligned}
Q_i = X W^Q_i&, \quad
K_i = X W^K_i, \quad
V_i = X W^V_i, \\
W^Q_i&,\, W^K_i,\, W^V_i \in \mathbb{R}^{d \times d_h},
\end{aligned}
\end{equation*}
where \(d_h = d / H\) is the dimensionality of each head, and \(H\) is the total number of heads. The attention weights and head outputs are then given by:
\begin{equation*}
\small
A_i = \mathrm{softmax}\left(\frac{Q_i K_i^\top}{\sqrt{d_h}}\right),\quad Z_i = A_i V_i\in \mathbb{R}^{n \times d_h}.
\end{equation*}
Let \(Z = \text{Concat}[Z_1, \dots, Z_H] \in \mathbb{R}^{n \times d}\) denote the concatenated outputs of all attention heads~\citep{nanda2022transformerlens}.  
The final \textbf{attention output} is obtained by applying the output projection:
\begin{equation*}
\small
\begin{aligned}
O &= Z W^O
   = [Z_1, \dots, Z_H]\,[W^O_1,\dots,W^O_H]^\top \\
  &= \sum_{i=1}^{H} Z_i W^O_i
   = \sum_{i=1}^{H} O_i ,
\end{aligned}
\end{equation*}
where each \(W^O_i \in \mathbb{R}^{d_h \times d}\) is the corresponding submatrix of \(W^O \in \mathbb{R}^{d \times d}\) associated with each head \(i\).

This formulation makes explicit that \(O\) is the sum of the outputs from all heads, where each head produces a rank-\(d_h\) output that is projected into the residual stream space through its corresponding \(W^O_h\). Thus, \(O\) represents the attention block’s total contribution to the residual stream.

\subsection{TopK Sparse Autoencoders}

In this work, we adopt the TopK sparse autoencoder (TopK SAE) introduced by \citet{gao2024oaisae}. Unlike standard SAEs that impose an \(\ell_1\) penalty, TopK SAE enforces exact sparsity by keeping only the top-\(k\) activations in the latent representation for each input. Formally, given an input vector \(x \in \mathbb{R}^d\), the encoder produces
\begin{equation*}
\small
z = \mathrm{TopK}(W_e x + b_e),
\end{equation*}
where \(\mathrm{TopK}(v)\) sets to zero all but the largest \(k\) entries of \(v\). The decoder then reconstructs
\begin{equation*}
\small
\hat{x} = W_d z + b_d.
\end{equation*}
The model is trained to minimize the reconstruction loss, optionally augmented with an auxiliary loss to prevent dead latents:
\begin{equation*}
\small
\mathcal{L}_{\text{TopK-SAE}} = \|x - \hat{x}\|_2^2 + \alpha \cdot \mathcal{L}_{\text{aux}},
\end{equation*}
where \(\mathcal{L}_{\text{aux}}\) is an optional term designed to penalize latents that never activate over a training period, and \(\alpha\) balances reconstruction fidelity and latent utilization.

\section{Low-Rank Structure of Attention Outputs}
\label{sec:activation_spectra}

We begin by presenting our central empirical finding: in Transformer models, attention outputs consistently display the strongest low-rank structure compared to MLP outputs and residual streams. As shown in Figure~\ref{fig:effective_rank}, attention outputs have a significantly lower effective rank. This phenomenon is remarkably robust, holding across different layers, model families and datasets. Further details regarding the activation sources are provided in Section~\ref{sec:activation_spectra:settings} and Appendix~\ref{appendix:model_dataset}. These observations highlight that the attention block modifies a subspace of the residual stream, while the MLP operates nearly on the full space.

\subsection{Quantifying Low-Rankness with Effective Rank}
\label{sec:activation_spectra:svd}

We consider the activation matrix $A \in \mathbb{R}^{n \times d}$, where each row corresponds to the activation vector of a single token. Here, $n$ denotes the number of data points and $d$ the dimensionality of the activation space (e.g., the model's hidden size). Unless otherwise specified, $\widetilde{A}$ represents mean-centered activations.

To quantify the effective dimensionality of the activations, we adopt the \textbf{\emph{effective rank}} metric introduced by~\citet{roy2007erank}. 

\begin{definition}[Effective Rank,~\citet{roy2007erank}]
Let $\widetilde{A}$ be a nonzero matrix with singular value decomposition $\widetilde{A} = U \Sigma V^\top$, where $\Sigma = \mathrm{diag}(\sigma_1, \sigma_2, \ldots, \sigma_r)$ contains singular values in descending order. Define the normalized singular value distribution as
\begin{equation*}
\small
p_k = \frac{\sigma_k}{\sum_{j=1}^{r} \sigma_j}, \quad k = 1, 2, \ldots, r.
\end{equation*}
The (Shannon) entropy of this distribution is
\begin{equation*}
\small
H(p_1, p_2, \ldots, p_r) = - \sum_{k=1}^{r} p_k \log p_k.
\end{equation*}
Then, the \textit{effective rank} of $\widetilde{A}$ is defined as
\begin{equation*}
\small
\mathrm{erank}(\widetilde{A}) = \exp\!\big\{ H(p_1, p_2, \ldots, p_r) \big\}.
\end{equation*}
\end{definition}

Intuitively, the effective rank captures how evenly the singular values are distributed—higher values indicate a more isotropic spectrum, whereas lower values reflect concentration along a few dominant directions. \textit{Fraction of effective rank} used in Figure~\ref{fig:effective_rank} means effective rank divided by the dimension of activation space.

Following \citet{bricken2023monosemanticity} and \citet{rajamanoharan2025gatedsae}, we compute the \textbf{\emph{fraction of downstream loss recovered}} by varying the number of retained components. We decompose the activations into singular value components and take the language model loss under full activation ablation as a baseline. As components are progressively reintroduced, we report the fraction of this ablated loss that is recovered. See Appendix~\ref{appendix:formal_fraction_loss_recovered} for a formal definition.

These quantitative measures complement our core analyses by providing a numerical characterization of the low-rank structure present in activations.

\subsection{Experiment Settings}
\label{sec:activation_spectra:settings}

For each activation type, we collect a total of 10 million activation vectors.\footnote{In extremely rare cases, outlier activations inflate variance along certain directions, potentially biasing variance-based dimensionality estimates. To mitigate this, we exclude activations whose norms exceed $5\sigma$ from the mean.} We empirically verified this sample size suffices to ensure stable and reproducible singular spectrum analyses in Appendix~\ref{appendix:error_analysis_svd}.

Unless otherwise specified, all experiments in Section~\ref{sec:activation_spectra} run on the middle layer of Llama-3.1-8B~(layer 15, zero-indexed), using SlimPajama dataset.

\subsection{Empirical Evidence of Low-Rank Structure}
\label{sec:activation_spectra:empirical}

\begin{figure}[htbp]
    \centering
    \begin{subfigure}[b]{0.23\textwidth}
        \centering
        \includegraphics[width=\textwidth]{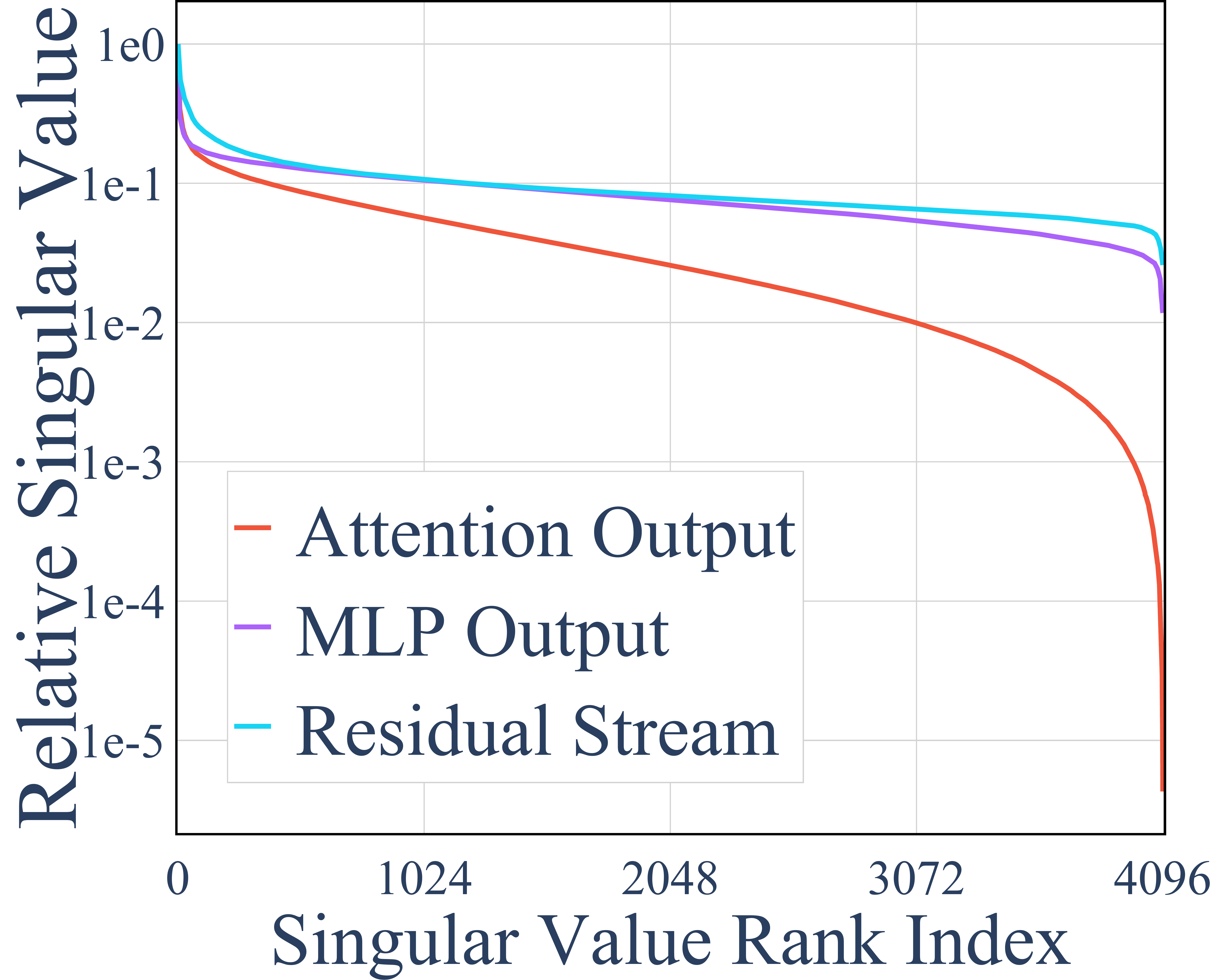}
        \caption{Singular Value Spectra.}
        \label{fig:rsv_spectra}
    \end{subfigure}
    \begin{subfigure}[b]{0.23\textwidth}
        \centering
        \includegraphics[width=\textwidth]{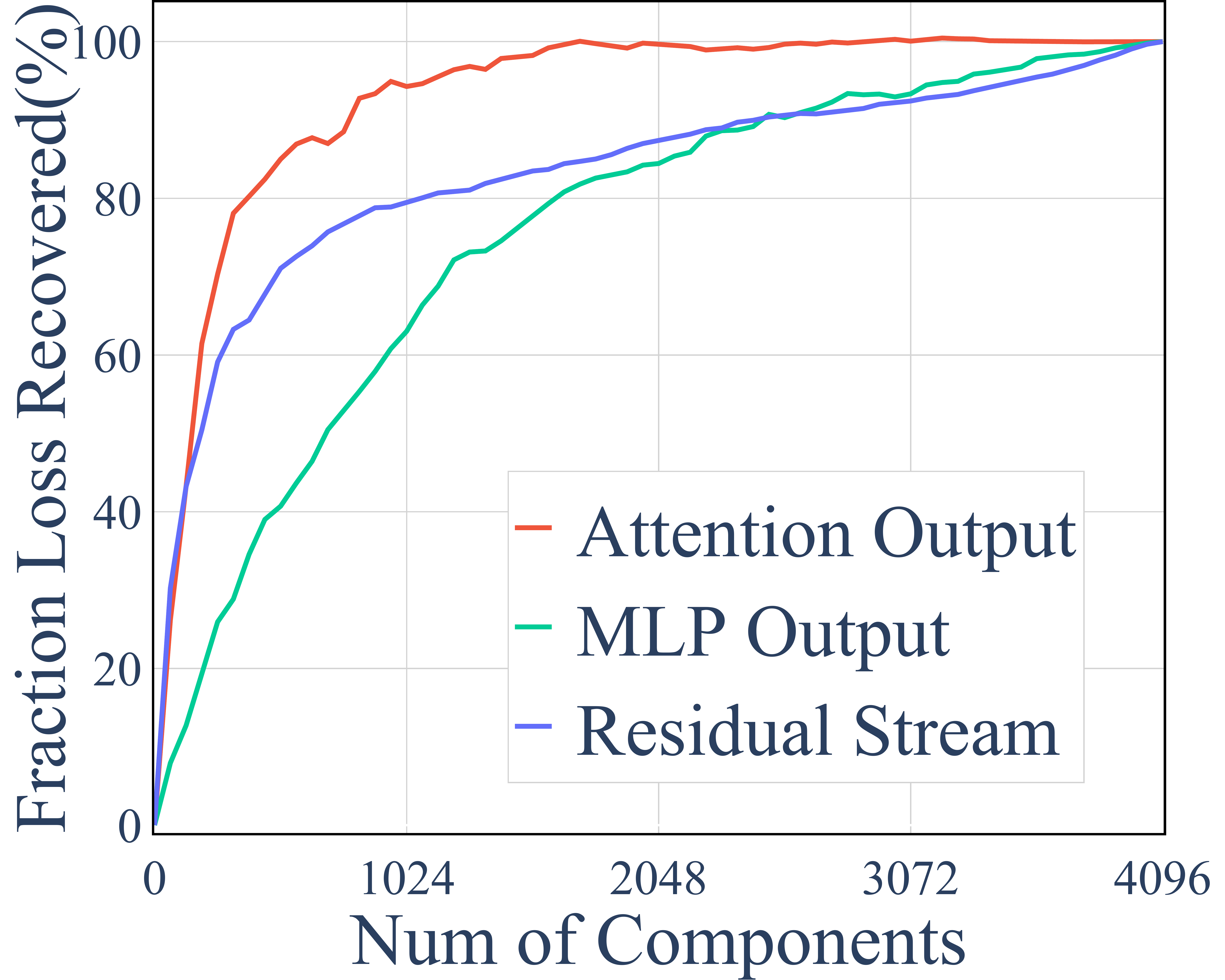}
        \caption{Fraction of Loss Recovered.}
        \label{fig:dim_ablate_downstream_loss}
    \end{subfigure}
    \caption{
    \textbf{(a)} The attention output is the most low-rank, as indicated by the sharpest decay in singular values. \textbf{(b)} Fraction of loss recovered using varying numbers of top singular components. 
    }
    \label{fig:rsv_delta_lm_loss}
\end{figure}

We draw our findings from three lines of evidence:
\paragraph{Low Effective Rank of Attention Output}
Attention outputs have a effective rank of around 60\% of the total dimensionality. In contrast, MLP outputs and the residual streams show much higher effective rank around 90\%~(Figure~\ref{fig:effective_rank}).

\paragraph{Rapid Singular Spectral Decay in Attention Output}
This is quantitatively evidenced by the number of components retaining significant energy: only 74.7\% singular values exceed 1\%\footnote{The resolution of bfloat16~\citep{kalamkar2019bf16} is 0.01.} of the maximum in attention output, versus 100.0\% for MLP output and residual stream~(Figure~\ref{fig:rsv_spectra}).

\paragraph{Efficient Downstream Loss Recovery}
Compared to zero ablation, attention output requires only 39.1\% of dimensions to recover over 99\% of the downstream loss, versus 95.3\% and 96.9\% of the dimensions for MLP outputs and residual streams to recover the same proportion~(Figure~\ref{fig:dim_ablate_downstream_loss}).

More results of these metrics across different layers, models, datasets and activation positions are shown in Appendix~\ref{appendix:more_low_rank_result}.

\subsection{The Output Projection Matrix Further Reduces the Effective Rank of Attention Outputs}
\label{sec:activation_spectra:origin}

Among all activation types, attention outputs consistently exhibits the most rapid singular spectral decay. To investigate whether this low-rank structure originates from the attention heads outputs~(\(Z\)), the output projection matrix (\(W^O\)), or their interaction, we perform a decomposition-based analysis.

Recall that the attention output is computed as \(O = Z W^O\), where \(Z \in \mathbb{R}^{n \times d}\) is the concatenated output of all attention heads, and \(W^O \in \mathbb{R}^{d \times d}\) is a learned linear projection. To understand how the singular value spectra of $O$ is shaped, we analyze the variance\footnote{For zero-mean activations, singular values correspond to the standard deviations of activations along the associated singular directions.} of \(O\) along a unit direction \(\hat{e} \in \mathbb{R}^d\), given by:
\begin{equation*}
\small
\mathrm{Var}(O\hat{e}) = \mathrm{Var}(Z W^O \hat{e}).
\end{equation*}
This expression highlights that the variance along \(e\) is determined by two factors: the norm of \(W^O \hat{e}\) and the variance of \(Z\) along \(W^O \hat{e}\). Specifically, we can rewrite the variance as:
\begin{equation*}
\small
\mathrm{Var}(O\hat{e}) = \mathrm{Var}(Z \hat{v}) \cdot \|v\|_2^2 \; \text{where} \; v = W^O \hat{e},\ \hat{v} = \frac{v}{\|v\|_2}.
\end{equation*}
We refer to \(\mathrm{Var}(Z \hat{v})\) as the \textbf{contribution of $Z$}, capturing how much variance the head output \(Z\) provides in that direction, and \(\|v\|_2^2\) as the \textbf{contribution of $W^O$}, measuring how much the output projection \(W^O\) scales or suppresses that direction.

\begin{figure}[t]
    \centering
    \includegraphics[width=0.44\textwidth]{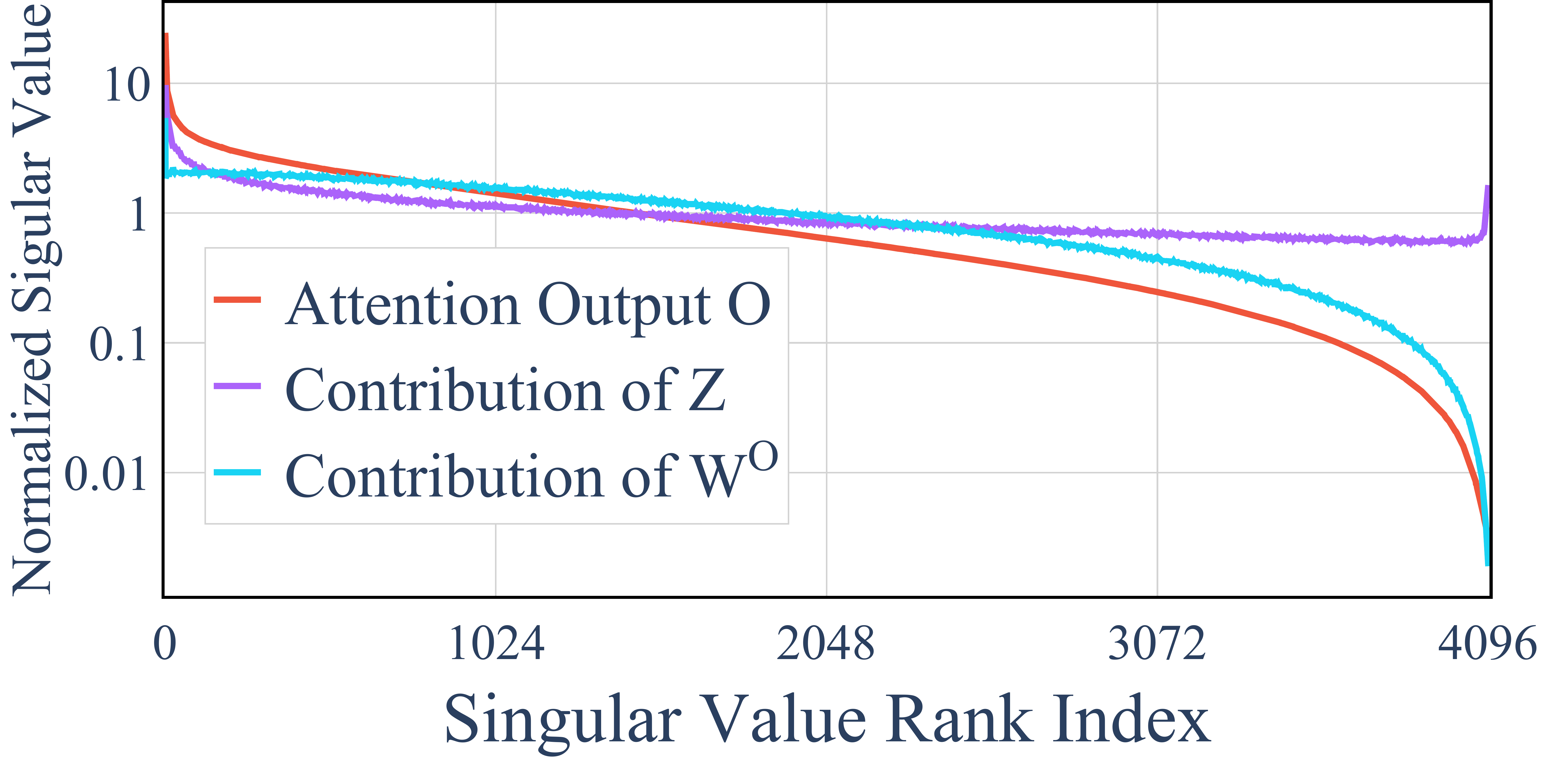}
    \caption{
        Decomposition of singular value spectra in \textcolor[HTML]{ef593f}{attention output $O$}. 
        We analyze the contributions of the \textcolor[HTML]{b270fa}{\textit{concatenated head outputs} $Z$} and the \textcolor[HTML]{2ad6f4}{\textit{projection matrix} $W^O$} to the singular value of \textcolor[HTML]{ef593f}{$O$}~($=$\textcolor[HTML]{b270fa}{$Z$}\textcolor[HTML]{2ad6f4}{$W^O$}). 
        For each component, the \textcolor[HTML]{ef593f}{red} value is the product of the \textcolor[HTML]{b270fa}{purple} and \textcolor[HTML]{2ad6f4}{blue} values. 
        The curve of \textcolor[HTML]{ef593f}{$O$} closely follow that of \textcolor[HTML]{b270fa}{$Z$} for the top components, whereas its downward trend at the tail is mainly due to \textcolor[HTML]{2ad6f4}{$W^O$ contribution}.
    }
    \label{fig:z_wo}
\end{figure}

We compute and visualize the singular values of attention output $O$ and these two contributions, as shown in Figure~\ref{fig:z_wo}. This analysis reveals that the low-rank structure of attention outputs is strongly influenced by \(W^O\), which further compresses $Z$ into a lower-dimensional subspace. The analysis of the effective rank of $Z$ in Figure~\ref{fig:erank_all_type} further supports this conclusion. From a mechanistic perspective, an intuitive explanation is that although each attention head contributes a $d_{\text{head}}$-dimensional subspace, the superposition of attention heads~\citep{jermyn2024attentionsuperposition, he2025lorsa} inevitably induces overlaps among these subspaces. 
Let $O_i$ denote the output of the $i^{\text{th}}$ attention head. Consequently, the dimension of the MHSA output satisfies

\begin{equation*}
\small
\begin{aligned}
\dim\!\left(\bigcup_i \text{span}(O_i)\right)
&\;\leq\; \sum_i \dim\!\big(\text{span}(O_i)\big) \\
&\;=\; d_{\text{head}} \cdot n_{\text{head}} \\
&\;=\; d_{\text{model}} \quad \text{(in standard MHSA)} .
\end{aligned}
\end{equation*}

\begin{figure*}[t]
    \centering
    \includegraphics[width=\textwidth]{pics/dead_erank_combined.pdf}
    \caption{
    The number of dead features~(left) and the effective rank~(mid) of each activation in Llama-3.1-8B, shows a surprising consistency~(right): activations with lower effective rank have more dead features, corresponding to all layers of attention output and last two layers of MLP output.
    }
    \label{fig:dead_erank}
\end{figure*}

\begin{figure*}[t]
    \centering
    \includegraphics[width=\textwidth]{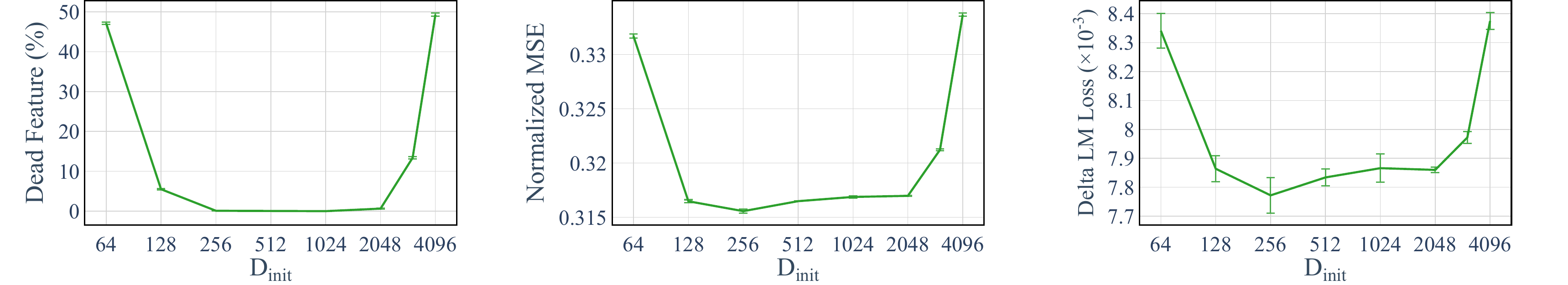}
    \caption{After using ASI, proportion of dead features~(left), normalized MSE~(mid) and Delta LM loss~(right) across different $d_{init}$ for activations with a full space dimension of 4096. All experiments repeat 3 times using different random seeds and error bars indicate mean ± std.}
    \label{fig:d_init}
\end{figure*}

\begin{figure*}[t]
    \centering
    \includegraphics[width=\textwidth]{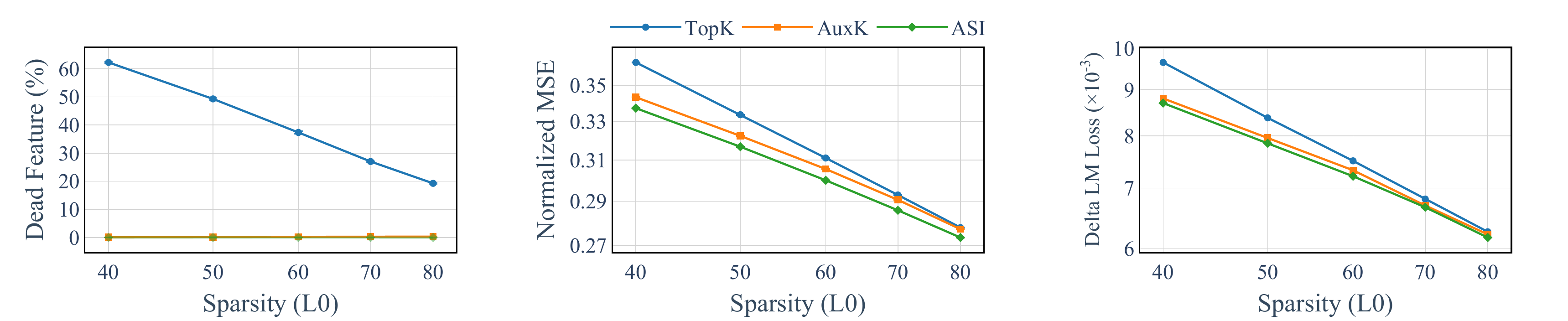}
    \caption{At a fixed number of features ($n = 32768$), \textcolor[HTML]{2ca02c}{ASI}~(TopK SAE with Active Subspace Init) achieves a better reconstruction-sparsity trade-off than \textcolor[HTML]{1f77b4}{TopK}~(standard TopK SAE) and \textcolor[HTML]{ff7f0e}{AuxK}~(TopK SAE with auxiliary loss). A similar trend is observed in its impact on Delta LM Loss. All experiments repeat 3 times using different random seeds and show the mean. The results of std are in Appendix~\ref{appendix:complete_results_metrics-l0} due to resolution constraints.
    }
    \label{fig:mse-delta_lm_loss-l0}
\end{figure*}

\section{Active Subspace Initialization for Sparse Autoencoders}
\label{sec:leveraging_subspace_improve_sae}

\subsection{Empirical Correlation Between Low-Rank Structure and Dead Features}
\label{sec:leveraging_subspace_improve_sae:low_rank_dead_cooccur}

To study how low-rankness affects the interpretability of attention, we adopt the same framework and dataset as the original LlamaScope study~\citep{he2024llamascope} to evaluate their SAEs trained on attention outputs, MLP outputs, and the residual stream\footnote{Another prominent set of open-source SAEs, GemmaScope~\citep{lieberum2024gemmascope}, train their attention SAEs on Z rather than attention output.}. We find that the number of dead features is strongly related to effective rank, as shown in Figure~\ref{fig:dead_erank}. This observation suggests that dead features may stem from the low-rank geometry of the activation space. We also train SAEs using different SAE hyperparameters and further systematically verify that this phenomenon is prevalent in Appendix~\ref{appendix:dead_features_effective_rank}.

\begin{figure*}[h]
    \begin{subfigure}[h]{0.99\textwidth}
        \centering
        \includegraphics[width=\textwidth]{pics/legend_only.pdf}
    \end{subfigure}
    \centering
    \begin{subfigure}[h]{0.31\textwidth}
        \centering
        \includegraphics[width=\textwidth]{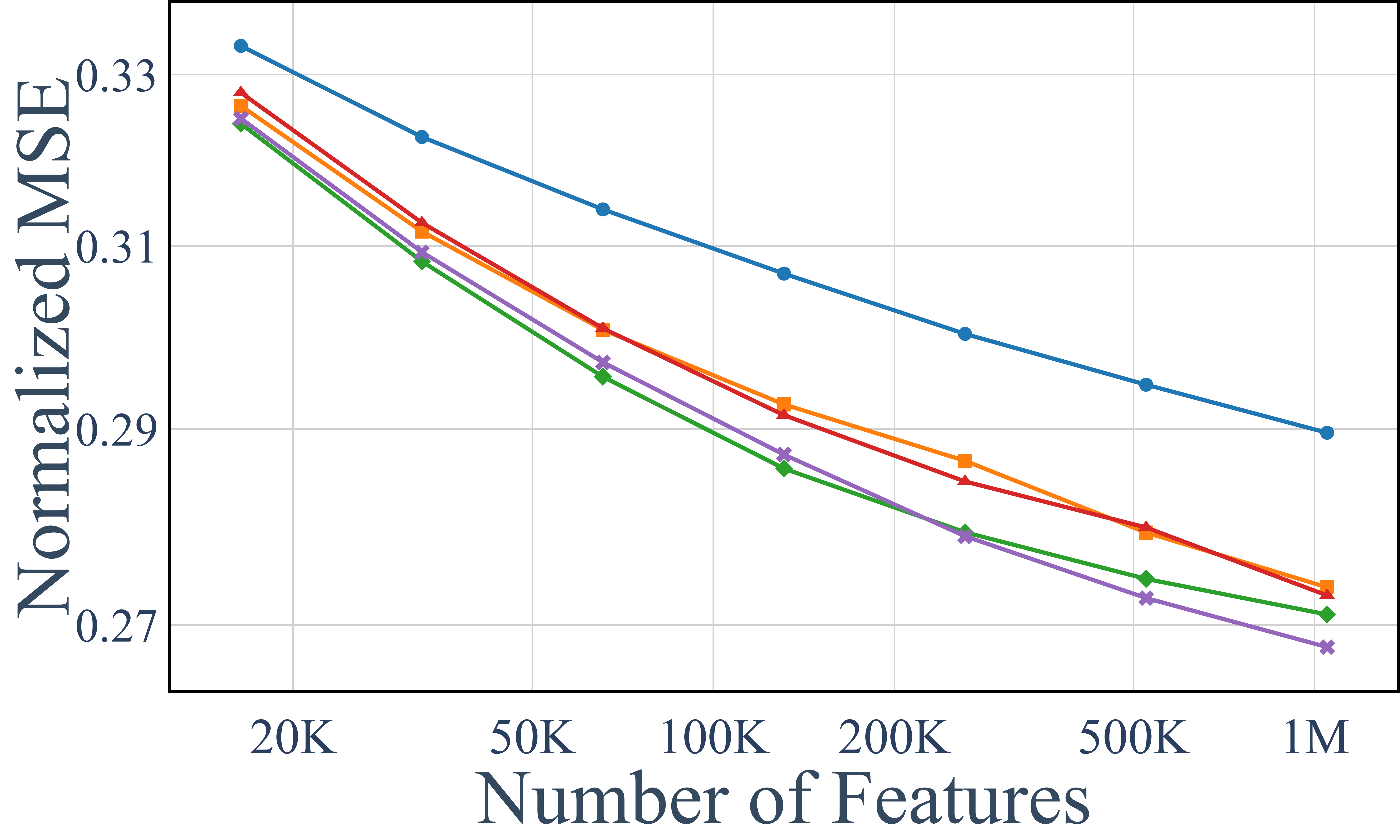}
        \caption{Loss vs. \#Total Features.}
        \label{fig:nmse_vs_features}
    \end{subfigure}
    \hfill
    \begin{subfigure}[h]{0.31\textwidth}
        \centering
        \includegraphics[width=\textwidth]{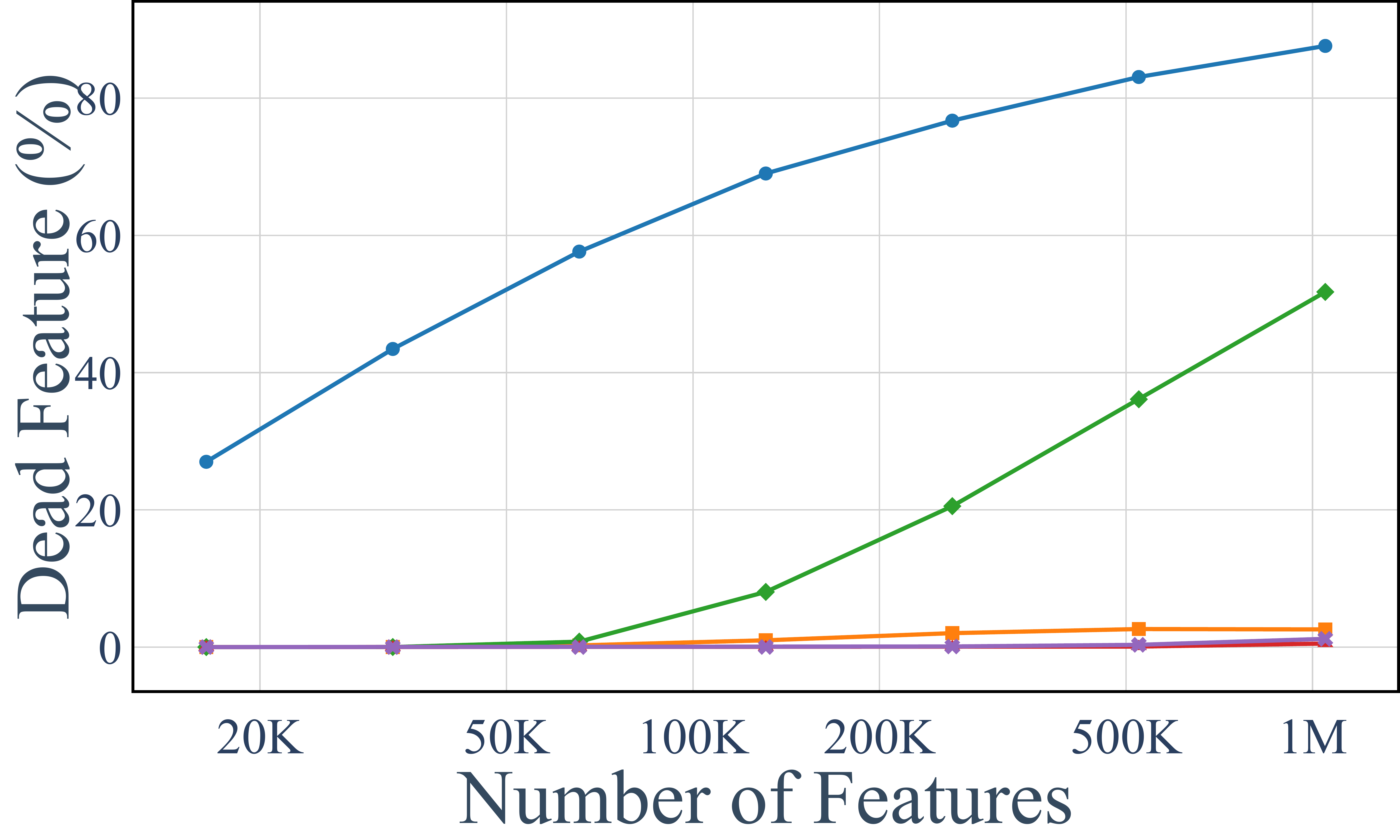}
        \caption{\#Dead Features vs. \#Total Features}
        \label{fig:dead_vs_features}
    \end{subfigure}
    \hfill
    \begin{subfigure}[h]{0.31\textwidth}
        \centering
        \includegraphics[width=\textwidth]{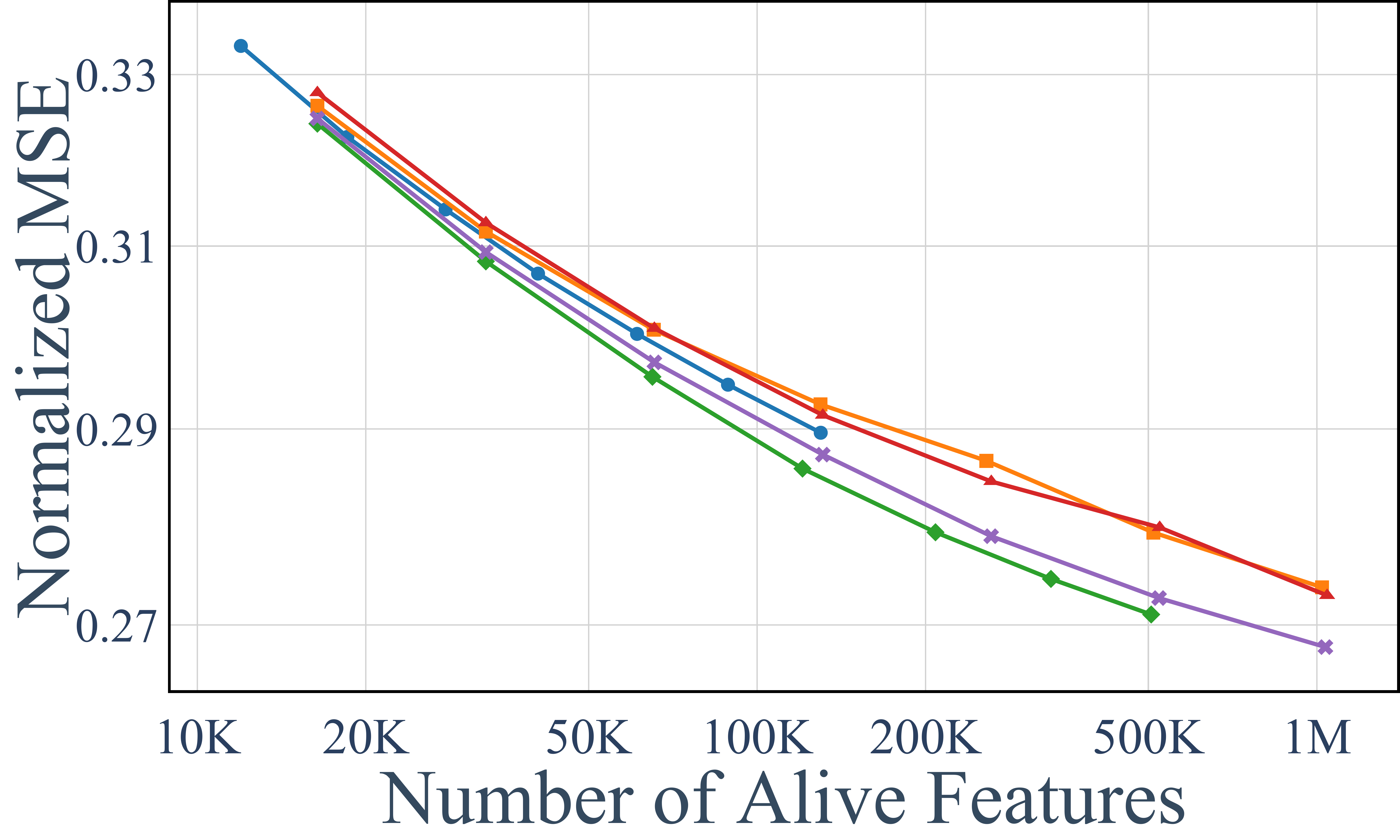}
        \caption{Loss vs. \#Alive Features.}
        \label{fig:nmse_vs_alive_features}
    \end{subfigure}
    \caption{Scaling results of TopK SAEs and their variants enhanced with \textit{AuxK}, \textit{Active Subspace Init}, and \textit{SparseAdam}--all trained on attention output from the middle layer of Llama-3.1-8B.
    (A) Loss at convergence across different feature counts: \textcolor[HTML]{2ca02c}{Active Subspace Init} consistently achieves lower reconstruction error than \textcolor[HTML]{1f77b4}{TopK} and \textcolor[HTML]{ff7f0e}{AuxK}. \textcolor[HTML]{9467bd}{Active Subspace Init with SparseAdam} achieves the best at large scale.
    (B) Dead features: \textcolor[HTML]{2ca02c}{Active Subspace Init} reduces dead features compared to \textcolor[HTML]{1f77b4}{TopK}, but still retains many at extremely large scales. \textcolor[HTML]{9467bd}{Enhanced with SparseAdam}, dead features can be reduced to less than 1\%.
    (C) Loss across different number of \textbf{alive} features: \textcolor[HTML]{2ca02c}{Active Subspace Init} achieves the most efficient utilization of \textbf{alive} features, while \textcolor[HTML]{ff7f0e}{AuxK} shows the lowest efficiency. Details in Section~\ref{sec:leveraging_subspace_improve_sae:scaling_law}.}
    \label{fig:scaling_law}
\end{figure*}

\subsection{Active Subspace Initialization for Sparse Autoencoders}
\label{sec:leveraging_subspace_improve_sae:subspace_init}


Based on this observation, we propose \textit{Active Subspace Initialization} (ASI), a lightweight and generalizable strategy for scaling SAEs to high capacities. 
Let $d$ denote the input dimension, $h$ the hidden dimension of the SAE, and $n$ the number of data points. 
Given activation matrices $\widetilde{A} \in \mathbb{R}^{n \times d}$ with singular value decomposition $\widetilde{A} = U \Sigma V^\top$ and $V \in \mathbb{R}^{d \times d}$ contains the right singular vectors, we select the top $d_{\text{init}}$ singular vectors to define the active subspace:
\begin{equation*}
\small
V_{\text{active}} = V_{:,:d_{\text{init}}} \in \mathbb{R}^{d \times d_{\text{init}}}.
\end{equation*}
To initialize the SAE within this subspace, we first randomly initialize the decoder weights $W_D \in \mathbb{R}^{h \times d}$ and then \emph{project their first $d_{\text{init}}$ columns onto the active subspace}:
\begin{equation*}
\small
W_D \leftarrow W_D \, V_{\text{active}} \, V_{\text{active}}^\top,  \qquad   W_E = W_D^\top.
\end{equation*}
where $W_E$ is the encoder weight matrix and $W_D$ is the decoder weight matrix. 
Intuitively, ASI aligns the initial SAE parameters with the effective directions of the data, ensuring that SAEs start in a meaningful low-dimensional subspace. As $d_{\text{init}}$ decreases from the full space dimension\footnote{Setting $d_{\text{init}}$ equal to the full space dimension is equivalent to not using Active Subspace Initialization.} within a certain range, the number of dead features in the Attention Output SAE rapidly drops, with a corresponding improvement in Mean Square Error~(MSE) and Delta LM loss\footnote{Following \citet{he2024llamascope}, this metric is defined as the difference between the original language model loss and the loss when the SAE is inserted at the corresponding position, evaluated over 1 million tokens.}~(Figure~\ref{fig:d_init}). We refer readers to Appendix~\ref{appendix:sae_training_details} for full SAE training details. Additional ablations, including the random subspace initialization baseline and the effect of activation rank, are reported in Appendix~\ref{appendix:ablation_study}. Pseudocode for Active Subspace Initialization (ASI) is provided in Appendix~\ref{appendix:pseudo_code}.

Using \textbf{Active Subspace Initialization} offers several benefits:

\paragraph{Reduced Dead Features and Enhanced Sparsity-Reconstruction Frontier Without Additional Compute}
It achieves near-zero dead features and slightly superior results compared to the auxiliary loss approach~(\textcolor[HTML]{ff7f0e}{AuxK}), at no additional computational cost of the same order.~(Figure~\ref{fig:mse-delta_lm_loss-l0}). 

\paragraph{Optimal Scaling Characteristics}  
Our approach demonstrates optimal scaling behavior across various SAE training methods. It outperforms \textcolor[HTML]{1f77b4}{TopK} and \textcolor[HTML]{ff7f0e}{AuxK} in any evaluated scale, from $16$K to $1$M features~(Section~\ref{sec:leveraging_subspace_improve_sae:scaling_law}).

\paragraph{General Applicability}
The technique maintains applicability to diverse architectural variants and activation functions, as it operates directly on the intrinsic properties of activations. This generalizability is further explored in Section~\ref{sec:leveraging_subspace_improve_sae:general_to_lorsa} and Appendix~\ref{appeneix:asi_other_activation_fuction}.

We conduct a statistical significance test in Appendix~\ref{appendix:significance} to demonstrate the statistical significance of our conclusions. We validate the effectiveness of ASI across different layers, models, and datasets in Appendix~\ref{appendix:across_layer_model_dataset}. We further compare the features of \textcolor[HTML]{1f77b4}{TopK} and \textcolor[HTML]{2ca02c}{ASI} in Appendix~\ref{appendix:compare_feature}, analyzing both the degree of monosemanticity and the behavior of SAE features in the dead subspace, to ensure that ASI increases the number of alive features \textbf{while preserving feature quality and maintaining the dictionary's coverage and reconstruction performance in the dead subspace.}

\begin{figure*}[t]
    \centering
    \includegraphics[width=\textwidth]{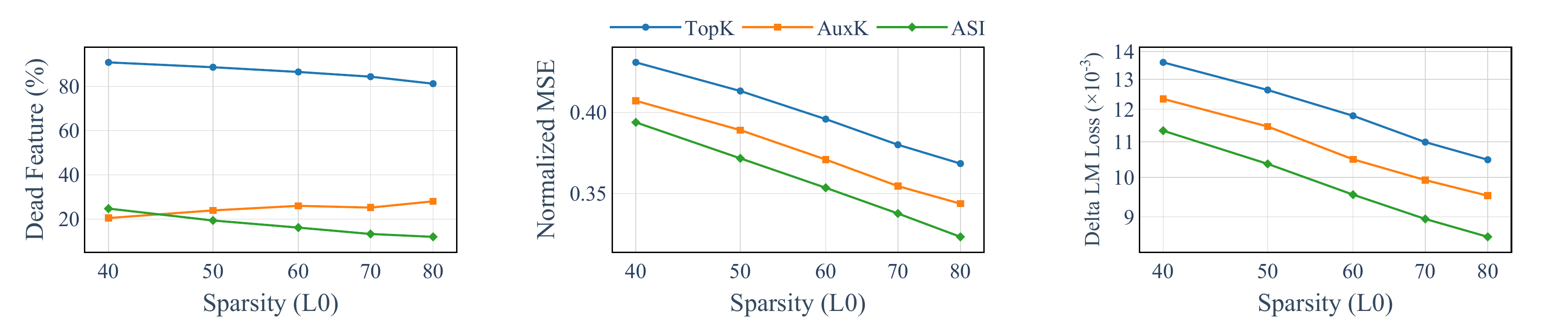}
    \caption{At a fixed number of Lorsa heads ($n = 32768$), \textcolor[HTML]{2ca02c}{ASI}~(TopK Lorsa with Active Subspace Init) achieves a better reconstruction-sparsity trade-off than \textcolor[HTML]{1f77b4}{TopK}~(standard TopK Lorsa) and \textcolor[HTML]{ff7f0e}{AuxK}~(TopK Lorsa with auxiliary loss). A similar trend is observed in its impact on Delta LM Loss.}
    \label{fig:lorsa_asi}
\end{figure*}

\subsection{Scaling Laws}
\label{sec:leveraging_subspace_improve_sae:scaling_law}

To assess scaling, we evaluate our method as the number of SAE features grows from $16$K to $1$M, keeping other hyperparameters fixed (see Appendix~\ref{appendix:sae_training_details}).

\paragraph{Active Subspace Init Improves Reconstruction.}  
As shown in Figure~\ref{fig:nmse_vs_features}, \textcolor[HTML]{2ca02c}{Active Subspace Init} consistently outperforms \textcolor[HTML]{1f77b4}{TopK} and \textcolor[HTML]{ff7f0e}{AuxK} across all scales.

\paragraph{Caveat: Some Dead Features Remain at Extremely Large Scales in Active Subspace Init.}  
Figure~\ref{fig:dead_vs_features} shows that, when scaling to extremely large feature counts, \textcolor[HTML]{2ca02c}{Active Subspace Init} produces more dead features than \textcolor[HTML]{ff7f0e}{AuxK}. However, reconstruction performance remains better, indicating that the revived features from \textcolor[HTML]{ff7f0e}{AuxK} contribute little to actual reconstruction quality (Figure~\ref{fig:nmse_vs_alive_features}).

\paragraph{Use Active Subspace Init with SparseAdam Further Improves Performance.}  
Prior work~\citep{brichen2023stale} identified \emph{stale momentum} as a key factor in dead feature formation. Building on this insight, we propose using \textbf{SparseAdam}, an optimizer specifically designed for sparse activation settings. By updating only the momentum terms and parameters corresponding to non-zero gradients, SparseAdam avoids stale momentum and thus mitigates the dead feature issue. As shown in Figures~\ref{fig:nmse_vs_features}, ~\ref{fig:dead_vs_features}, combining \textcolor[HTML]{9467bd}{Active Subspace Init with SparseAdam} substantially reduces dead features while reaching the lowest reconstruction error. While orthogonal to our initialization method, this choice provides a practical complement that further stabilizes training when scaling SAEs to very large capacities. We discuss more about \emph{stale momentum} and \textbf{SparseAdam} in Appendix~\ref{appendix:stale_momentum}.

\subsection{Generalizing to Sparse Replacement Models}
\label{sec:leveraging_subspace_improve_sae:general_to_lorsa}

Recent work by \citet{he2025lorsa} shows that Lorsa---a sparse replacement for MHSA---has a substantial fraction of dead parameters. 
We hypothesize this is partly due to initialization: standard random initialization ignore the low-dimensional active subspace of attention outputs (Section~\ref{sec:activation_spectra}).

\paragraph{Applying ASI to Lorsa.}
To test whether our subspace-based approach extends beyond SAEs, we incorporate ASI (Section~\ref{sec:leveraging_subspace_improve_sae:subspace_init}) into Lorsa training.
When training Lorsa to approximate a target MHSA, we partition Lorsa heads into such groups, matching the number of attention heads in the MHSA. For each group, we initialize the encoder and decoder matrices directly within the corresponding input and output active subspaces of the MHSA head it corresponds. In addition, we initialize each group's Q/K weights from the MHSA Q/K parameters. Pseudo code is provided in Appendix~\ref{appendix:pseudo_code}

\paragraph{Results.}
This initialization sharply reduces dead parameters under identical hyperparameters (Figure~\ref{fig:lorsa_asi}) while improving reconstruction quality. Further Lorsa training details are in Appendix~\ref{appendix:lorsa_details}.

\section{Discussion and Limitations}

\paragraph{Low-rank attention outputs suggest a new direction for model improvement.}
Modern large language models are predominantly built by stacking attention and MLP blocks with residual connections. Our finding that attention outputs show a low-rank structure suggests a potentially new avenue for improving model capacity and expressivity: future architectures may benefit from explicitly mitigating this rank collapse within attention modules. Notably, this phenomenon persists across models with and without \textit{grouped-query attention}~(GQA), and under both relative and absolute positional encodings~(Section~\ref{sec:activation_spectra}), indicating that it may reflect a more fundamental property of the attention mechanism itself rather than an artifact of specific design choices.

\paragraph{Causality between Low-Rank Structure and Dead Features.}
We find a strong correlation between low-rank activations and the emergence of dead features (Section~\ref{sec:leveraging_subspace_improve_sae}), but the underlying causal mechanism is unresolved. This effect may arise from optimization dynamics or feature competition, and we leave a rigorous explanation to future work.

\paragraph{When to Use Active Subspace Initialization.}
Active Subspace Initialization is most beneficial when activations exhibit pronounced low-rank structure, such as attention outputs, residual streams of some narrow datasets, or certain specialized situations. For activation sites without clear low-rank behavior, the improvements are marginal (Appendix~\ref{appendix:ablation_study:asi_on_resid}).

\section{Conclusion}
\label{sec:conclusion}

We identified the low-rank structure of attention outputs as a fundamental property of Transformer models and a key cause of dead features in sparse dictionary learning. Our proposed \textit{Active Subspace Initialization} method addresses this by aligning SAE features with the intrinsic geometry of activations, reducing dead features while improving reconstruction quality. The approach generalizes beyond SAEs to sparse replacement models.





\nocite{langley00}

\bibliography{example_paper}
\bibliographystyle{icml2026}

\newpage
\appendix
\onecolumn
\section*{Author Contributions}
\label{appendix:author_contribution}

\textbf{Junxuan Wang} and \textbf{Zhengfu He} co-discovered the low-rank structure in attention outputs and its correlation with dead features. 

\textbf{Junxuan Wang} proposed the Active Subspace Initialization method and the use of SparseAdam for SAE training, conducted all experiments.

\textbf{Xuyang Ge}, \textbf{Zhengfu He}, and \textbf{Wentao Shu} made core contributions to the SAE training codebase infrastructure. 

\textbf{Zhengfu He} helped to edit the manuscript.

\textbf{Xipeng Qiu} supervised the research and provided research guidance.

\section{Activation Sources}
\label{appendix:model_dataset}
The spectral characteristics of activations vary substantially across model architectures, datasets, and positional contexts. Below, we describe the experimental configurations used to support a broad and representative analysis.

\paragraph{Models}
We study four large language models of different families--\texttt{GPT-2}\footnote{\href{https://huggingface.co/openai-community/gpt2}{https://huggingface.co/openai-community/gpt2}}, \texttt{Llama-3.1-8B}\footnote{\href{https://huggingface.co/meta-llama/Llama-3.1-8B}{https://huggingface.co/meta-llama/Llama-3.1-8B}}, \texttt{Qwen3-8B}\footnote{\href{https://huggingface.co/Qwen/Qwen3-8B}{https://huggingface.co/Qwen/Qwen3-8B}}, and \texttt{Gemma-2-9B}\footnote{\href{https://huggingface.co/google/gemma-2-9b}{https://huggingface.co/google/gemma-2-9b}}--all based on the Transformer architecture. This allows us to assess the robustness of spectral properties under varying model training configurations.

\paragraph{Datasets}
To investigate how dataset diversity affects activation spectra, we select two datasets with varying linguistic and domain characteristics: (1) \texttt{SlimPajama}\footnote{\href{https://huggingface.co/datasets/cerebras/SlimPajama-627B}{https://huggingface.co/datasets/cerebras/SlimPajama-627B}}, an English corpus comprising web text, Github, Arxiv and other sources. \textit{Multi-Corpora} in Figure~\ref{fig:effective_rank} denotes the setting where all SlimPajama components are jointly used with random mixing. \textit{Github} and \textit{Arxiv} correspond to the code and scientific paper subsets of SlimPajama, respectively. (2) \texttt{CCI3-Data}\footnote{\href{https://huggingface.co/datasets/BAAI/CCI3-Data}{https://huggingface.co/datasets/BAAI/CCI3-Data}}, a Chinese dataset with broad domain coverage, which is used in Appendix~\ref{appendix:more_low_rank_result} as a supplement.

\paragraph{Activation Positions}
We analyze three types of activations: (1) attention output, (2) MLP output, and (3) residual stream~(post layer). 

\section{Formal Definition of \textit{Fraction of Loss Recovered}}
\label{appendix:formal_fraction_loss_recovered}

Given the original language model cross-entropy loss is \(loss_{\text{original}}\), the loss after ablating the activation at a specific position to zero is \(loss_{\text{zero}}\), and the loss after replacing the original activation projected to the subspace spanned by first $n$ singular vectors is \(loss_{\text{recovered}}\). Then, for these $n$ components, the fraction loss recovered is calculated as:  
\[
\frac{loss_{\text{zero}} - loss_{\text{recovered}}}{loss_{\text{zero}} - loss_{\text{original}}}
\]  

\section{Error analysis in Singular Value Decomposition}
\label{appendix:error_analysis_svd}

For the attention output of layer 15 of Llama-3.1-8B, we performed 5 times of singular value decompositions, using different 10 million tokens for each, and calculated the Coefficient of Variation (CV) for each singular value across these 5 runs. The maximum CV was only \(4.9 \times 10^{-3}\), and the mean and standard deviation of the effective rank computed from these 5 SVD results were 2523.165 and 0.404, respectively, with a CV of \(1.5 \times 10^{-4}\). These error experiments show that using 10 million tokens for singular value decomposition is sufficiently stable.

\section{More Singular Spectrum and Effective Rank Results}
\label{appendix:more_low_rank_result}

\begin{figure}[htbp]
    \centering

    \begin{subfigure}[t]{0.45\textwidth}
        \centering
        \includegraphics[width=\textwidth]{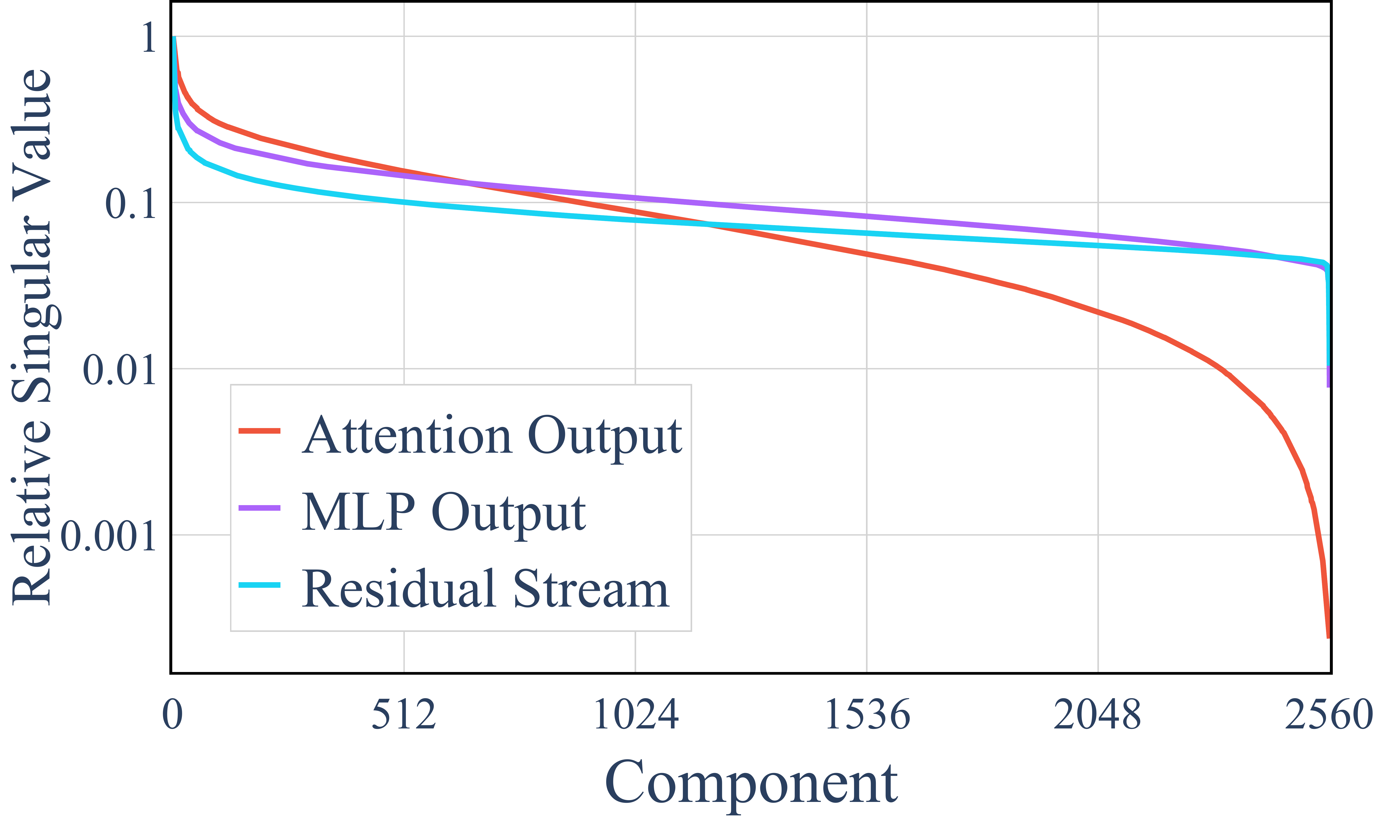}
        \caption{
            Middle layer of pythia-2.8b; SlimPajama \\
            Effective Dimensionality: Attention Output 1670; \\
            MLP Output 2252; Residual Stream 2327
        }
        \label{fig:rsv_pythia_2.8b_slimpajama}
    \end{subfigure}
    \hfill
    \begin{subfigure}[t]{0.45\textwidth}
        \centering
        \includegraphics[width=\textwidth]{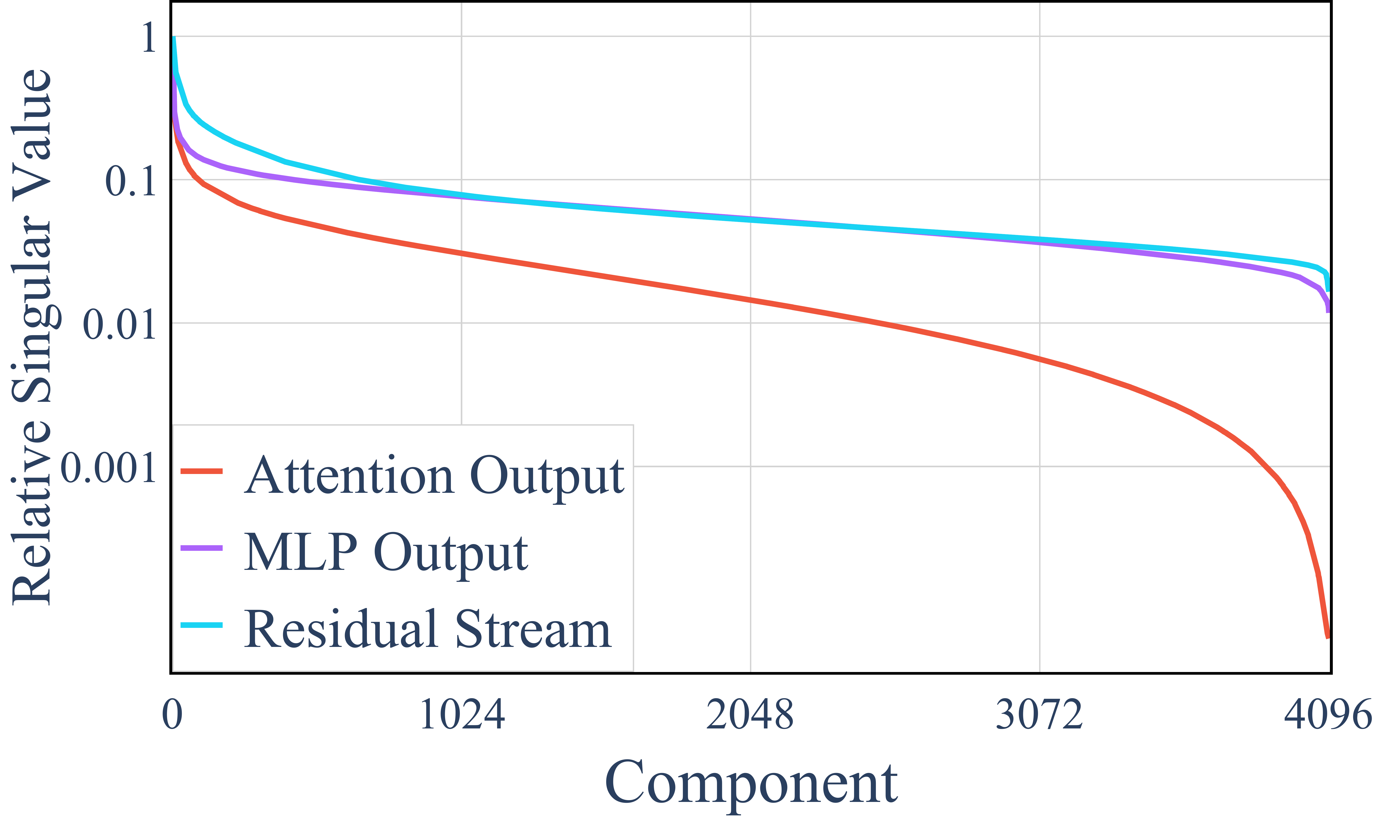}
        \caption{
            Middle layer of Qwen3-8B; CCI3-Data \\
            Effective Dimensionality: Attention Output 2356; \\
            MLP Output 3558; Residual Stream 3140
        }
        \label{fig:rsv_qwen3_8b_cci3}
    \end{subfigure}

    \begin{subfigure}[t]{0.45\textwidth}
        \centering
        \includegraphics[width=\textwidth]{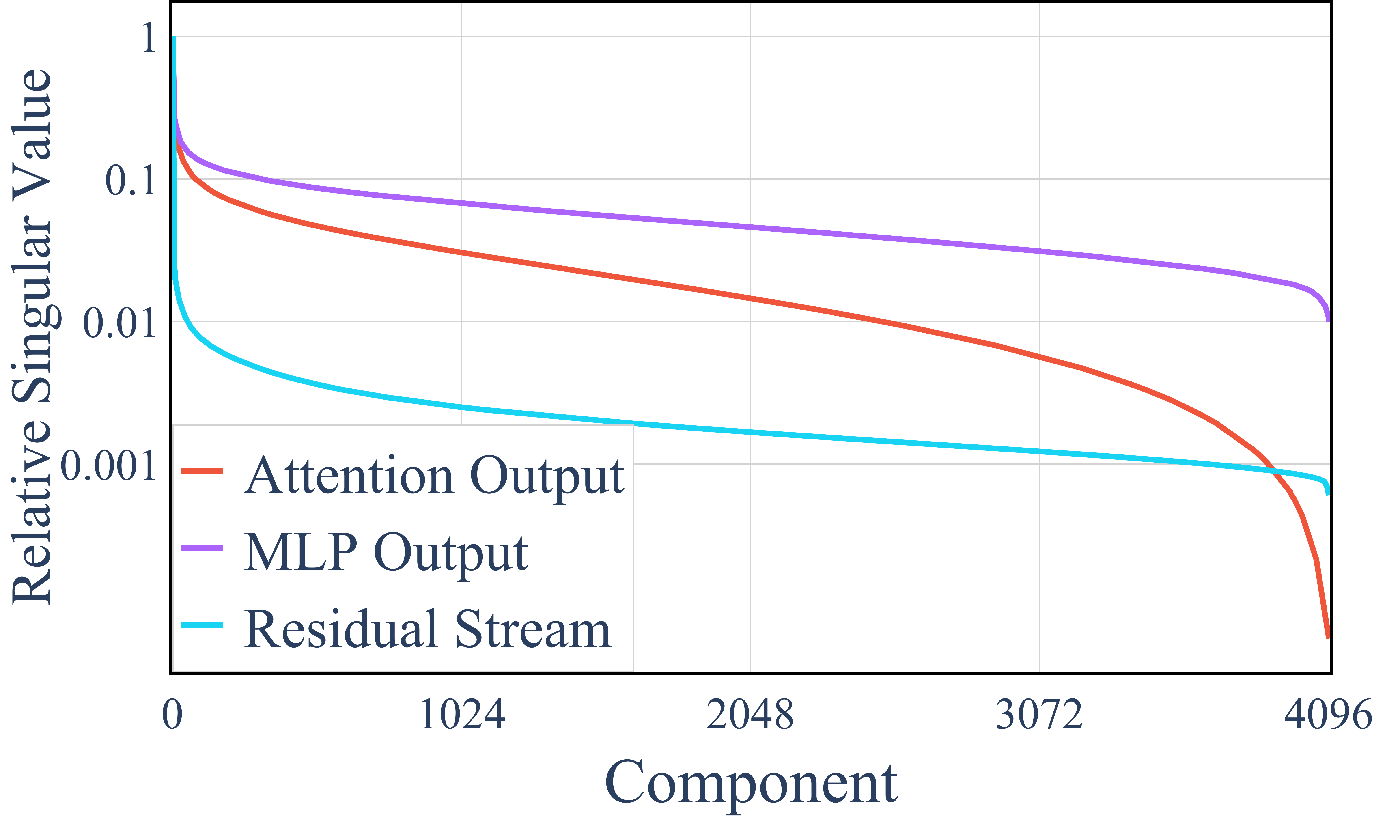}
        \caption{
            Middle layer of Qwen3-8B; SlimPajamaGithub \\
            Effective Dimensionality: Attention Output 2410; \\
            MLP Output 3495; Residual Stream 2000
        }
        \label{fig:rsv_qwen3_8b_github}
    \end{subfigure}
    \hfill
    \begin{subfigure}[t]{0.45\textwidth}
        \centering
        \includegraphics[width=\textwidth]{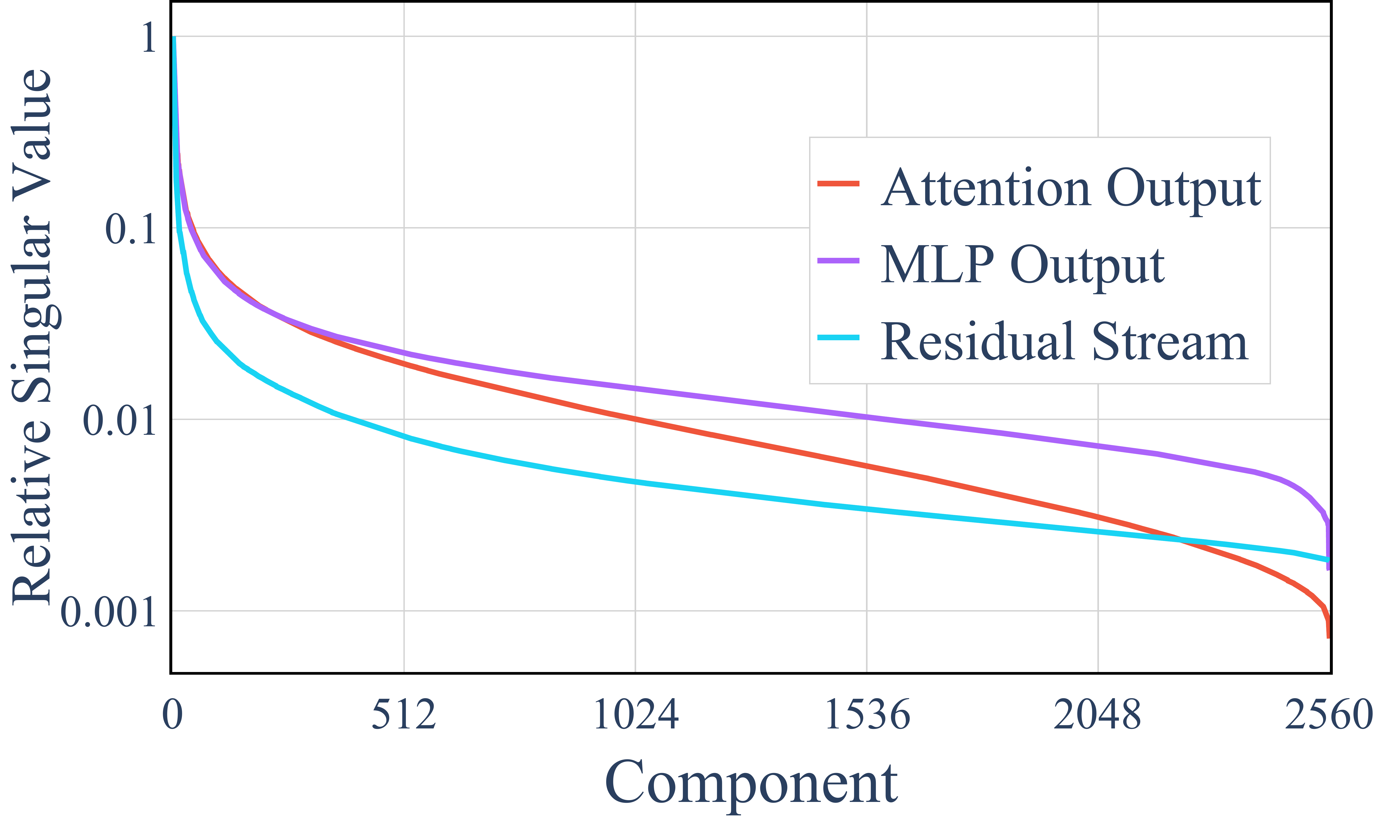}
        \caption{
            Middle layer of Qwen3-4B; SlimPajama \\
            Effective Dimensionality: Attention Output 1122; \\
            MLP Output 1485; Residual Stream 990
        }
        \label{fig:rsv_qwen3_4b_slimpajama}
    \end{subfigure}

    \begin{subfigure}[t]{0.45\textwidth}
        \centering
        \includegraphics[width=\textwidth]{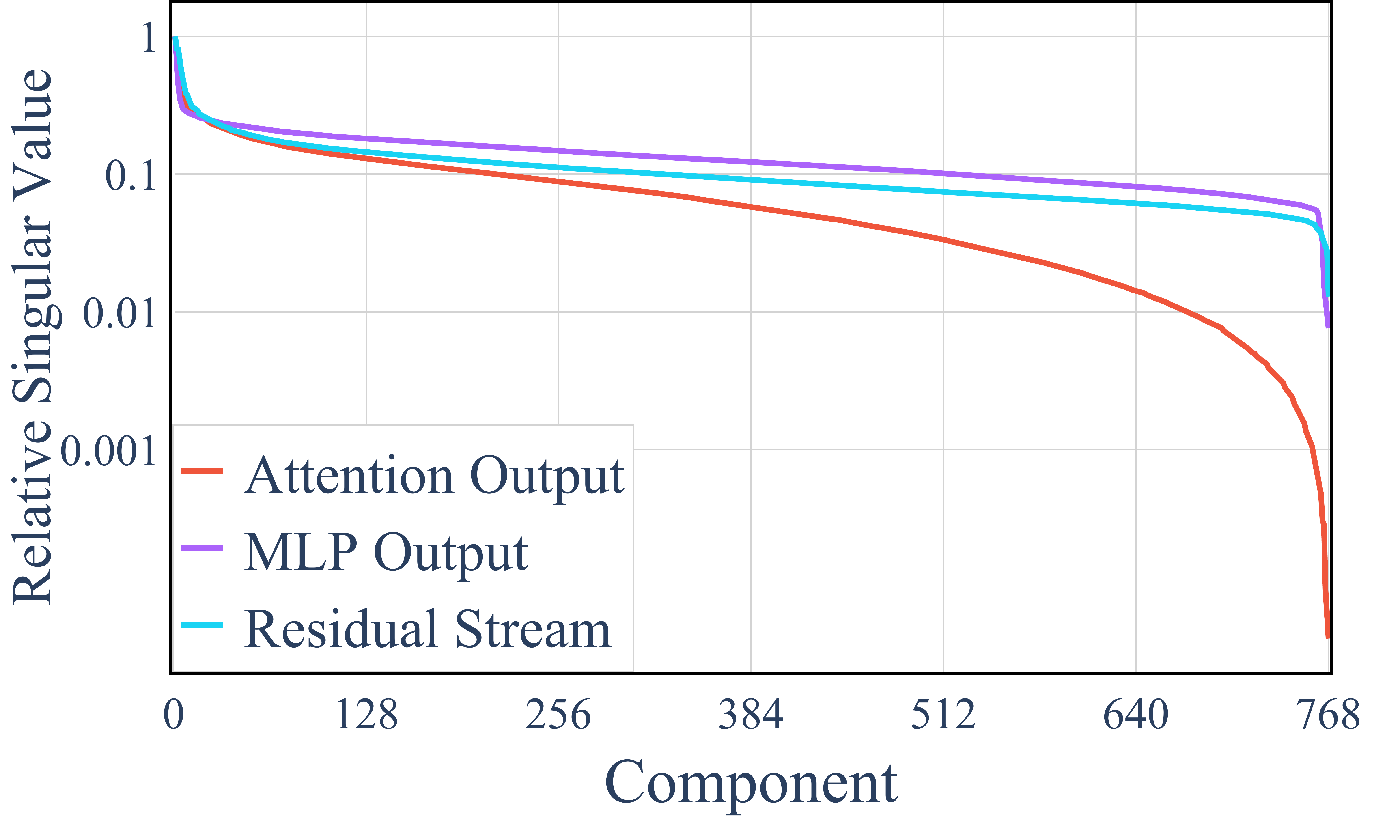}
        \caption{
            Middle layer of gpt2; SlimPajama \\
            Effective Dimensionality: Attention Output 515; \\
            MLP Output 703; Residual Stream 656
        }
        \label{fig:rsv_gpt2}
    \end{subfigure}
    \hfill
    \begin{subfigure}[t]{0.45\textwidth}
        \centering
        \includegraphics[width=\textwidth]{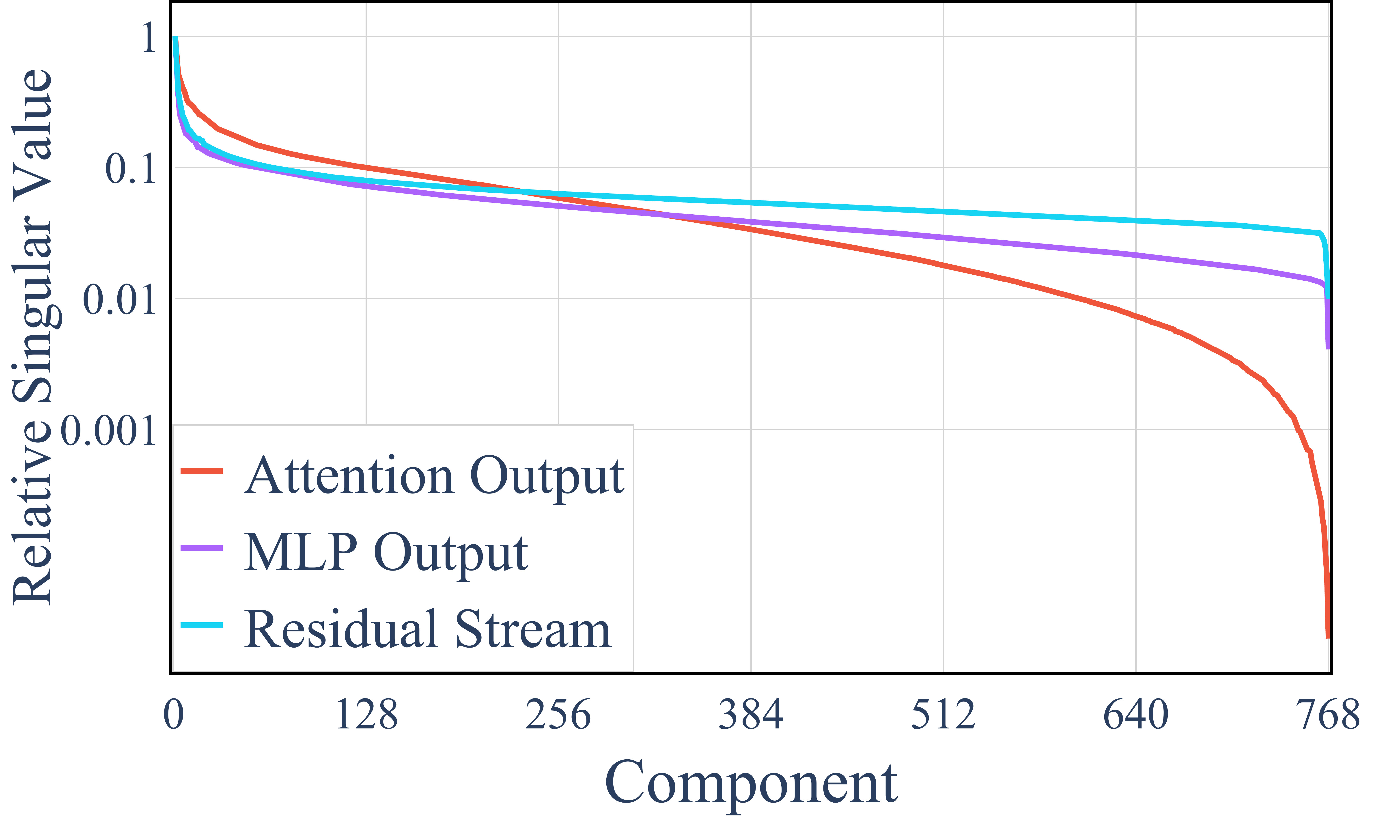}
        \caption{
            Middle layer of pythia-160m; SlimPajama \\
            Effective Dimensionality: Attention Output 434; \\
            MLP Output 594; Residual Stream 662
        }
        \label{fig:rsv_qwen3_4b_slimpajama}
    \end{subfigure}

    \caption{}
    \label{fig:rsv_appendix}
\end{figure}

\subsection{Across Models and Datasets}

We present relative singular values for some other model-dataset pairs in Figure~\ref{fig:rsv_appendix}. Models include 
pythia-160m\footnote{\href{https://huggingface.co/EleutherAI/pythia-160m}{https://huggingface.co/EleutherAI/pythia-160m}}, pythia-2.8b\footnote{\href{https://huggingface.co/EleutherAI/pythia-2.8b}{https://huggingface.co/EleutherAI/pythia-2.8b}}. Datasets include SlimPajamaGithub~(subset of SlimPajama) and CCI3-Data.

\subsection{Across Layers and Activation Positions}

\begin{figure}[htbp]
    \centering
    \begin{subfigure}[t]{0.7\textwidth}
        \centering
        \includegraphics[width=\textwidth]{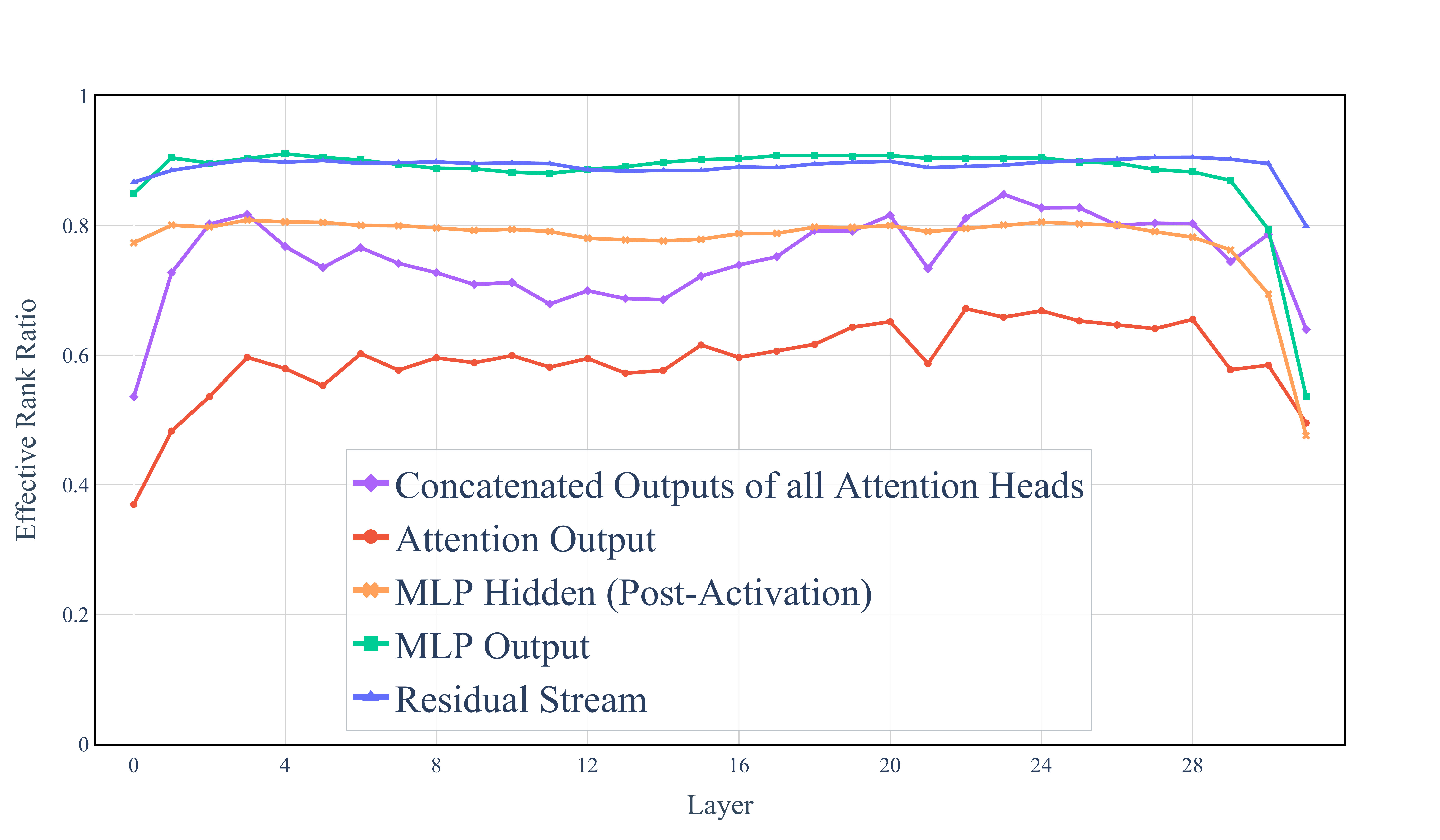}
        \caption{The number of the effective rank of each activation in Llama-3.1-8B}
        \label{fig:erank_all_type_llama}
    \end{subfigure}
    \centering
    \begin{subfigure}[t]{0.7\textwidth}
        \centering
        \includegraphics[width=\textwidth]{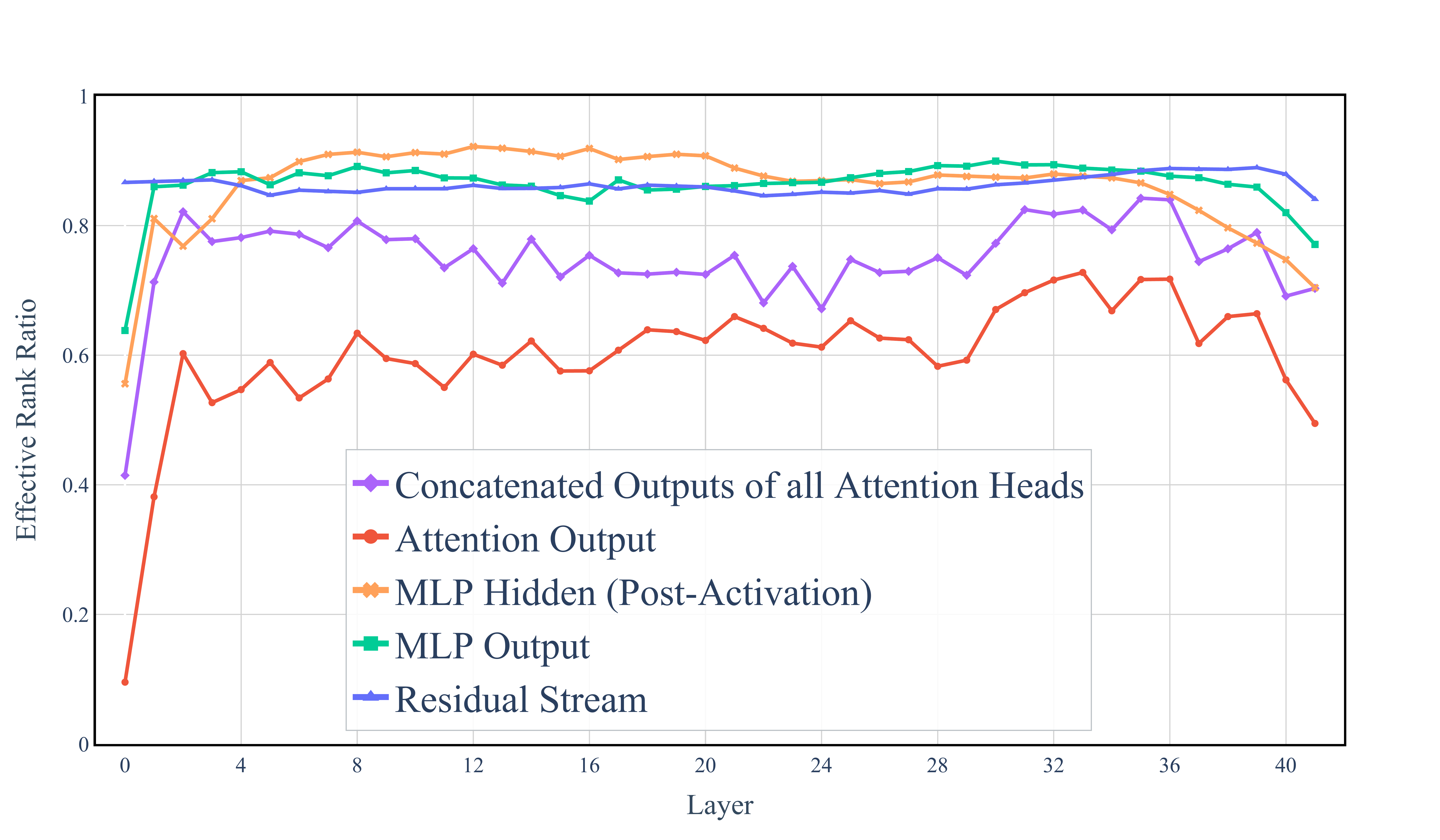}
        \caption{The number of the effective rank of each activation in Gemma-2-9B}
        \label{fig:erank_all_type_gemma}
    \end{subfigure}
    
    \centering
    \begin{subfigure}[t]{0.7\textwidth}
        \includegraphics[width=\textwidth]{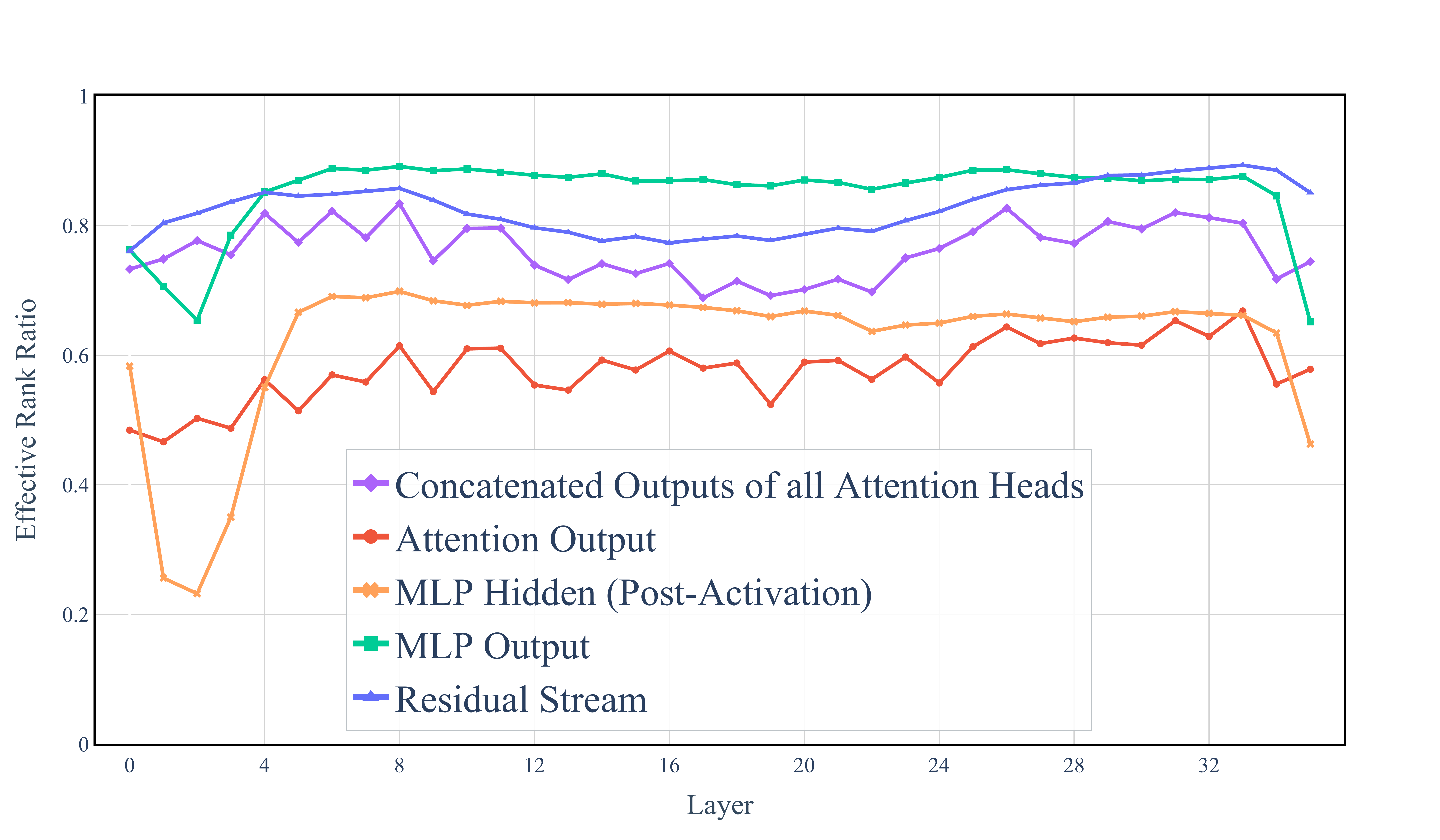}
        \caption{The number of the effective rank of each activation in Qwen3-8B}
        \label{fig:erank_all_type_qwen}
    \end{subfigure}
    
    \caption{}
    \label{fig:erank_all_type}
\end{figure}

We present the effective rank for activations that are commonly used to train SAEs in Figure~\ref{fig:erank_all_type}, including \textbf{Concatenated Outputs of all Attention Heads}~(Z), attention output, the hidden activations of MLP~(post activation fuction), MLP output, residual stream. All effective ranks are computed on SlimPajama.

\section{SAE Training Details}
\label{appendix:sae_training_details}

We train SAEs as the following description.

\subsection{Hyperparameters}
\label{appendix:sae_training_details:hyper}

\paragraph{Model, Dataset, Layer, Pos} Llama-3.1-8B, SlimPajama, 15(index start at 0), attention output.
\paragraph{Sparsity} We empirically set $k=50$ for a reasonable sparsity following~\citet{he2024llamascope}, except the experiments for sweeping $k$.
\paragraph{Dictionary Size} We empirically set $n_{features}=32768$ which is $8\times d_{model}$, except the experiments for sweeping dictionary size~(scaling law).
\paragraph{Batch Size} We empirically set the batch size to $4096$.
\paragraph{Optimizer} We use the Adam and SparseAdam optimizer, both with $\beta_1 = 0.9$, $\beta_2 = 0.999$, and $\epsilon = 10^{-8}$. Unless otherwise specified, Adam is used by default.
\paragraph{Learning Rate} The learning rate for \textbf{Adam} and \textbf{SparseAdam} is sweeped separately in [$1\mathrm{e}{-5}$, $2\mathrm{e}{-5}$, $4\mathrm{e}{-5}$, $6\mathrm{e}{-5}$, $8\mathrm{e}{-5}$, $1\mathrm{e}{-4}$, $2\mathrm{e}{-4}$, $4\mathrm{e}{-4}$], and we ultimately use $4\mathrm{e}{-5}$ for \textbf{Adam} and $6\mathrm{e}{-5}$ for \textbf{SparseAdam}. We employ a three-phase learning rate schedule consisting of a linear warm-up, a stable phase, and a linear decay. The learning rate increases linearly from zero to its maximum value over the first 500 steps, remains constant during the intermediate phase, and then decays linearly to 1\% of the maximum value over the final 20\% of the total training steps.

\paragraph{AuxK} We follow \citet{gao2024oaisae} to set auxiliary loss coefficient \(\alpha\) as $\frac{1}{32}$. We sweep the $k_{aux}$ in [256, 512, 1024, 2048] and finally choose 512. We also sweep \(\alpha\) and find the results are less sensitive to \(\alpha\) in a reasonable interval.
\paragraph{Dimension of Subspace for SAE Initialization~($d_{init}$)} We use 768 for all experiments, except the experiments for sweeping $d_{init}$. We refer readers to Appendix~\ref{appendix:ablation_study:d_init} for the reason.
\paragraph{Total Tokens} We use $800$M tokens for each SAE training.

\subsection{Collecting Activations}
We truncate each document to 1024 tokens and prepend a \textless bos\textgreater~token
to the beginning of each document.
During training, we exclude the activations corresponding to
the \textless bos\textgreater~, \textless eos\textgreater and \textless pad\textgreater~tokens.

It has been observed that activations from different sequence positions within the same document are often highly correlated and may lack diversity. To mitigate this issue, it is common to introduce randomness into the training data. Our shuffling strategy maintains a buffer that is reshuffled whenever the buffer is refilled.

\subsection{Initialization}

The decoder columns $W^{dec}_{:,i}$ are initialized uniformly, and the optimal norm for them is found through a grid search to minimize the initial reconstruction loss. We find that the specific initialization norm has little impact, as long as in a reasonable scope. For example, initializing $W^{dec}_{:,i}$ uniformly with a fixed bound, as in \citet{conerly2025dltechnique}, yields similar results. The encoder weights $W^{enc}$ are initialized as the transpose of $W^{dec}$,  while both the encoder bias $b^{enc}$ and decoder bias $b^{dec}$ are set to zero.

\subsection{Jumprelu SAEs}
\label{appendix:sae_training_details:jumprelu_sae}
We trained JumpReLU SAEs~\citep{rajamanoharan2024jumprelusae} under two distinct hyperparameter configurations: one maintaining a consistently low $\ell_0$ value throughout training, and another where $\ell_0$ is gradually decreased from a higher initial value. Unless otherwise specified, all JumpReLU SAEs were trained using the same settings as \citet{conerly2025dltechnique}, which corresponds to the latter configuration. The key modifications for the former setting are as follows: (1) we initialized the encoder bias to zero instead of applying the heuristic that equalizes feature activation counts at initialization, and (2) we kept the sparsity coefficient fixed rather than employing a global warm-up schedule. As a result, the $\ell_0$ sparsity level started at a relatively low value early in training. This design is critical to our approach: we observed that if the model remains in a high-$\ell_0$ regime (e.g., on the order of $d_{\text{model}} / 2$) for an extended period before sparsity increases, the feature directions tend to drift away from the active subspace during this phase, thereby diminishing the effectiveness of our method~(Appendix~\ref{appeneix:asi_other_activation_fuction:jumprelu}).

\section{Additional Analysis on Effective Dimensionality and Dead Features}
\label{appendix:dead_features_effective_rank}

This section provides additional experimental evidence supporting the claim that \emph{low effective dimensionality strongly correlates with a higher proportion of dead features} in sparse autoencoders (SAEs). In Figure~\ref{fig:dead_erank}, we report the effective rank of the residual stream, attention output, and MLP output at each layer of \texttt{Llama-3.1-8B}, along with the proportion of dead features in the SAEs trained on these activations. All SAEs were obtained from \texttt{Llamascope}~\citep{he2024llamascope}, which uses the same dictionary size (32768 features) and sparsity level ($L_0 = 50$).

To provide systematic analysis, we additionally train SAEs across multiple dictionary sizes (16384, 32768, 65536) and sparsity levels ($L_0 \in \{32, 64, 128\}$) following Appendix~\ref{appendix:sae_training_details} The SAEs are trained on the activations of \texttt{Llama-3.1-8B}, and the corresponding effective ranks of these activations can be found in Figure~4 of the main paper. Tables~\ref{tab:l0_32}--\ref{tab:l0_128} summarize the proportion of dead features across these configurations.

Across all settings, attention outputs consistently show substantially higher dead-feature ratios than residual streams. This trend holds even when the dictionary size varies by a factor of four and the sparsity level varies by a factor of four.

\begin{table}[h!]
    \centering
    \caption{Proportion of dead features for $L_0 = 32$ across different dictionary sizes.}
    \label{tab:l0_32}
    \begin{tabular}{lccc}
        \toprule
        \textbf{Activation (Effective Rank)} & \textbf{16384} & \textbf{32768} & \textbf{65536} \\
        \midrule
        Layer 7 attention (2351) & \textbf{84.80\%} & \textbf{90.31\%} & \textbf{94.02\%} \\
        Layer 7 residual (3664) & 1.61\% & 6.69\% & 17.10\% \\
        Layer 15 attention (2506) & \textbf{68.70\%} & \textbf{79.86\%} & \textbf{87.22\%} \\
        Layer 15 residual (3611) & 27.01\% & 45.70\% & 61.33\% \\
        Layer 23 attention (2654) & \textbf{58.45\%} & \textbf{70.48\%} & \textbf{78.81\%} \\
        Layer 23 residual (3634) & 0.13\% & 0.26\% & 1.35\% \\
        \bottomrule
    \end{tabular}
\end{table}

\begin{table}[h!]
    \centering
    \caption{Proportion of dead features for $L_0 = 64$ across different dictionary sizes.}
    \label{tab:l0_64}
    \begin{tabular}{lccc}
        \toprule
        \textbf{Activation (Effective Rank)} & \textbf{16384} & \textbf{32768} & \textbf{65536} \\
        \midrule
        Layer 7 attention (2351) & \textbf{66.97\%} & \textbf{75.65\%} & \textbf{82.96\%} \\
        Layer 7 residual (3664) & 0.02\% & 0.06\% & 0.15\% \\
        Layer 15 attention (2506) & \textbf{41.65\%} & \textbf{54.68\%} & \textbf{67.43\%} \\
        Layer 15 residual (3611) & 1.73\% & 7.96\% & 18.83\% \\
        Layer 23 attention (2654) & \textbf{41.58\%} & \textbf{56.70\%} & \textbf{67.05\%} \\
        Layer 23 residual (3634) & 0.15\% & 0.09\% & 0.08\% \\
        \bottomrule
    \end{tabular}
\end{table}

\begin{table}[h!]
    \centering
    \caption{Proportion of dead features for $L_0 = 128$ across different dictionary sizes.}
    \label{tab:l0_128}
    \begin{tabular}{lccc}
        \toprule
        \textbf{Activation (Effective Rank)} & \textbf{16384} & \textbf{32768} & \textbf{65536} \\
        \midrule
        Layer 7 attention (2351) & \textbf{49.85\%} & \textbf{56.96\%} & \textbf{64.83\%} \\
        Layer 7 residual (3664) & 0.00\% & 0.00\% & 0.02\% \\
        Layer 15 attention (2506) & \textbf{15.09\%} & \textbf{25.61\%} & \textbf{37.11\%} \\
        Layer 15 residual (3611) & 0.07\% & 0.12\% & 0.44\% \\
        Layer 23 attention (2654) & \textbf{21.90\%} & \textbf{39.02\%} & \textbf{52.68\%} \\
        Layer 23 residual (3634) & 0.14\% & 0.09\% & 0.07\% \\
        \bottomrule
    \end{tabular}
\end{table}

\section{Complete Results of SAE metrics}
\label{appendix:complete_results_metrics-l0}
We use 3 different random seeds for all experiments in Figure~\ref{fig:mse-delta_lm_loss-l0} and compute the mean values and the standard deviations of each metrics, as shown in Table~\ref{tab:mse-delta_lm_loss-l0}.

\begin{table}[t]
    \centering
    \caption{Comparison of Base, AuxK, and ASI across different $L_0$ settings. Numbers show mean $\pm$ std over 3 seeds.}
    \label{tab:mse-delta_lm_loss-l0}
    \small
    \begin{tabular}{c c c c c}
        \toprule
        $L_0$ & Metric & Base & AuxK & ASI \\
        \midrule
        \multirow{3}{*}{40}
            & Dead Feature & 20395.00 $\pm$ 72.77 & 37.00 $\pm$ 5.20 & \textbf{10.67 $\pm$ 1.15} \\
            & Normalized MSE & 0.36323 $\pm$ 0.00018 & 0.34328 $\pm$ 0.00011 & \textbf{0.33724 $\pm$ 0.00022} \\
            & Delta LM Loss ($\times 10^{-3}$) & 9.650 $\pm$ 0.040 & 8.801 $\pm$ 0.053 & \textbf{8.697 $\pm$ 0.033} \\
        \midrule
        \multirow{3}{*}{50}
            & Dead Feature & 16144.33 $\pm$ 129.52 & 54.67 $\pm$ 5.51 & \textbf{4.00 $\pm$ 1.73} \\
            & Normalized MSE & 0.33367 $\pm$ 0.00014 & 0.32241 $\pm$ 0.00020 & \textbf{0.31680 $\pm$ 0.00006} \\
            & Delta LM Loss ($\times 10^{-3}$) & 8.375 $\pm$ 0.029 & 7.958 $\pm$ 0.032 & \textbf{7.847 $\pm$ 0.050} \\
        \midrule
        \multirow{3}{*}{60}
            & Dead Feature & 12239.33 $\pm$ 165.51 & 76.00 $\pm$ 9.17 & \textbf{2.67 $\pm$ 1.53} \\
            & Normalized MSE & 0.31106 $\pm$ 0.00007 & 0.30555 $\pm$ 0.00007 & \textbf{0.30000 $\pm$ 0.00005} \\
            & Delta LM Loss ($\times 10^{-3}$) & 7.503 $\pm$ 0.059 & 7.325 $\pm$ 0.017 & \textbf{7.214 $\pm$ 0.026} \\
        \midrule
        \multirow{3}{*}{70}
            & Dead Feature & 8854.67 $\pm$ 40.51 & 115.33 $\pm$ 15.37 & \textbf{1.67 $\pm$ 0.58} \\
            & Normalized MSE & 0.29295 $\pm$ 0.00008 & 0.29064 $\pm$ 0.00008 & \textbf{0.28575 $\pm$ 0.00013} \\
            & Delta LM Loss ($\times 10^{-3}$) & 6.805 $\pm$ 0.074 & 6.694 $\pm$ 0.021 & \textbf{6.664 $\pm$ 0.066} \\
        \midrule
        \multirow{3}{*}{80}
            & Dead Feature & 6311.33 $\pm$ 55.77 & 109.67 $\pm$ 11.50 & \textbf{1.00 $\pm$ 1.73} \\
            & Normalized MSE & 0.27787 $\pm$ 0.00003 & 0.27715 $\pm$ 0.00003 & \textbf{0.27341 $\pm$ 0.00003} \\
            & Delta LM Loss ($\times 10^{-3}$) & 6.259 $\pm$ 0.005 & 6.217 $\pm$ 0.028 & \textbf{6.168 $\pm$ 0.033} \\
        \bottomrule
    \end{tabular}
\end{table}

\section{Statistical Significance Test}
\label{appendix:significance}

To assess whether the performance improvements introduced by ASI are statistically significant, we conducted a comprehensive significance analysis across multiple evaluation metrics. 

\subsection{Experimental Setup}

We evaluate the statistical significance of performance differences between ASI and two baseline methods (TopK and AuxK) under the following controlled setting:

\begin{itemize}
    \item \textbf{Model / Layer / $L_0$ / Dictionary Size:} Llama-3.1-8B, Layer 15, $L_0 = 50$, dictionary size = 32{,}768.
    \item \textbf{Number of runs:} 15 independent trials for each method, each with a different random seed.
    \item \textbf{Evaluation metrics:} Dead Feature Count, Normalized MSE, and $\Delta$ LM Loss. All metrics follow a ``lower is better'' criterion.
    \item \textbf{Comparisons performed:} ASI vs.\ TopK and ASI vs.\ AuxK for all three metrics.
\end{itemize}

\subsection{Hypothesis Testing Framework}

For each metric and each baseline method, we perform Welch's t-test (also known as Welch's unequal-variance t-test), which does not assume equal variances between groups.  
For ASI and a given baseline method, we test the following hypotheses:

\[
\begin{aligned}
    H_0 &: \mu_{\text{ASI}} \ge \mu_{\text{baseline}}
    && \text{(ASI is worse than or equal to the baseline)}, \\
    H_1 &: \mu_{\text{ASI}} < \mu_{\text{baseline}}
    && \text{(ASI outperforms the baseline)}.
\end{aligned}
\]

This is a one-tailed test, as we explicitly test whether ASI achieves significantly lower metric values.

We use the following SciPy function for all tests:
\[
\texttt{scipy.stats.ttest\_ind}(\text{ASI}, \text{baseline}, \texttt{equal\_var=False}, \texttt{alternative='less'}).
\]

\subsection{Results}

Table~\ref{tab:significance} reports the resulting p-values for all comparisons.  
A smaller p-value indicates stronger evidence that ASI outperforms the baseline.

\begin{table}[h]
    \centering
    \caption{Welch's t-test results for ASI compared with TopK and AuxK across 15 random seeds.}
    \label{tab:significance}
    \small
    \begin{tabular}{l l c}
        \toprule
        \textbf{Comparison} & \textbf{Metric} & \textbf{p-value} \\
        \midrule
        \multirow{3}{*}{ASI vs.\ TopK}
            & Dead Feature Count & $3.26 \times 10^{-35}$ \\
            & Normalized MSE     & $2.99 \times 10^{-40}$ \\
            & $\Delta$ LM Loss   & $1.14 \times 10^{-23}$ \\
        \midrule
        \multirow{3}{*}{ASI vs.\ AuxK}
            & Dead Feature Count & $4.33 \times 10^{-15}$ \\
            & Normalized MSE     & $6.33 \times 10^{-40}$ \\
            & $\Delta$ LM Loss   & $1.27 \times 10^{-6}$ \\
        \bottomrule
    \end{tabular}
\end{table}

Across all evaluation metrics and both baseline methods, the p-values are far below standard significance thresholds (e.g., $\alpha = 0.05$). Therefore, we reject the null hypothesis for all comparisons. These results demonstrate that the improvements achieved by ASI are statistically significant and robust across random seeds.

\section{Additional Evaluation Across Layers, Models, and Datasets}
\label{appendix:across_layer_model_dataset}

To assess the robustness and generality of ASI, we extend our experiments beyond the primary configuration used in the main paper (Llama-3.1-8B, Layer 15, SlimPajama). In particular, we investigate whether the advantages of ASI over baseline approaches (TopK and AuxK) persist across different layers, models, and datasets. We consider this evaluation essential, as mechanisms in sparse autoencoding can vary substantially across architectural depth, data distribution, and model family.

\subsection{Evaluation on Llama-3.1-8B Across Multiple Layers}
We first evaluate ASI on two additional layers of Llama-3.1-8B (Layers 7 and 23), using the SlimPajama dataset. Activations are taken from the attention output. Results are summarized in Table~\ref{tab:llama_layers}.

\begin{table}[h]
\centering
\caption{Performance of ASI and baseline methods on Llama-3.1-8B Layers 7 and 23. Lower values indicate better performance.}
\label{tab:llama_layers}
\begin{tabular}{lccccccccc}
\toprule
& \multicolumn{3}{c}{Dead Feature Count} 
& \multicolumn{3}{c}{Normalized MSE} 
& \multicolumn{3}{c}{$\Delta$ LM Loss ($\times 10^{-3}$)} \\
\cmidrule(lr){2-4} \cmidrule(lr){5-7} \cmidrule(lr){8-10}
Layer & Base & AuxK & ASI & Base & AuxK & ASI & Base & AuxK & ASI \\
\midrule
7  & 25836 & \textbf{98}  & 5308 & 0.32870 & 0.29800 & \textbf{0.28882} & 5.414 & 4.589 & \textbf{4.430} \\
23 & 15542 & 332 & \textbf{27}   & 0.21302 & 0.20235 & \textbf{0.20141} & 1.942 & 1.865 & \textbf{1.845} \\
\bottomrule
\end{tabular}
\end{table}

We observe that ASI consistently achieves the lowest reconstruction error (Normalized MSE) and the smallest degradation in language modeling performance ($\Delta$ LM loss). For Layer 7, ASI still retains a number of dead features, which may be attributed to its smaller effective rank (2351) compared to Layer 23 (2654). Since we use a fixed $d_{\text{init}}$ across layers, this mismatch can lead to remaining dead features. Despite this, ASI still achieves the lowest MSE on both layers.

\subsection{Evaluation on Qwen3-8B Across Layers and a New Dataset}
To further test cross-model and cross-dataset generality, we conduct experiments on the Qwen3-8B model using the fineweb-edu dataset, again using attention-output activations. Results for Layers 8 and 26 are shown in Table~\ref{tab:qwen_layers}.

\begin{table}[h]
\centering
\caption{Performance comparison on Qwen3-8B Layers 8 and 26 using the fineweb-edu dataset. Lower values indicate better performance.}
\label{tab:qwen_layers}
\begin{tabular}{lccccccccc}
\toprule
& \multicolumn{3}{c}{Dead Feature Count} 
& \multicolumn{3}{c}{Normalized MSE} 
& \multicolumn{3}{c}{$\Delta$ LM Loss ($\times 10^{-3}$)} \\
\cmidrule(lr){2-4} \cmidrule(lr){5-7} \cmidrule(lr){8-10}
Layer & Base & AuxK & ASI & Base & AuxK & ASI & Base & AuxK & ASI \\
\midrule
8  & 19048 & \textbf{4}   & 7   & 0.31228 & 0.28849 & \textbf{0.28533} 
   & 2.9561 & 2.5911 & \textbf{2.5667} \\
26 & 16286 & \textbf{66}  & 566 & 0.30090 & 0.28127 & \textbf{0.28087}
   & 1.3406 & 1.2980 & \textbf{1.2922} \\
\bottomrule
\end{tabular}
\end{table}

The results again confirm that ASI achieves the lowest reconstruction error and the smallest increase in LM loss across both layers. The near-dead-feature-free representation produced by AuxK is also observed, but ASI consistently outperforms AuxK in reconstruction quality and LM preservation.

\subsection{Summary}
Across all tested configurations––spanning multiple layers, two large language model families, and two datasets––ASI exhibits consistent advantages over baseline methods. These evaluations provide strong empirical evidence that the benefits of ASI are not confined to a specific layer, model, or dataset, but instead generalize across diverse settings.

\section{Ablation Study}
\label{appendix:ablation_study}

\subsection{Active Subspace Init vs Random Subspace Init}

\begin{figure}[htbp]
    \centering
    \includegraphics[width=\textwidth]{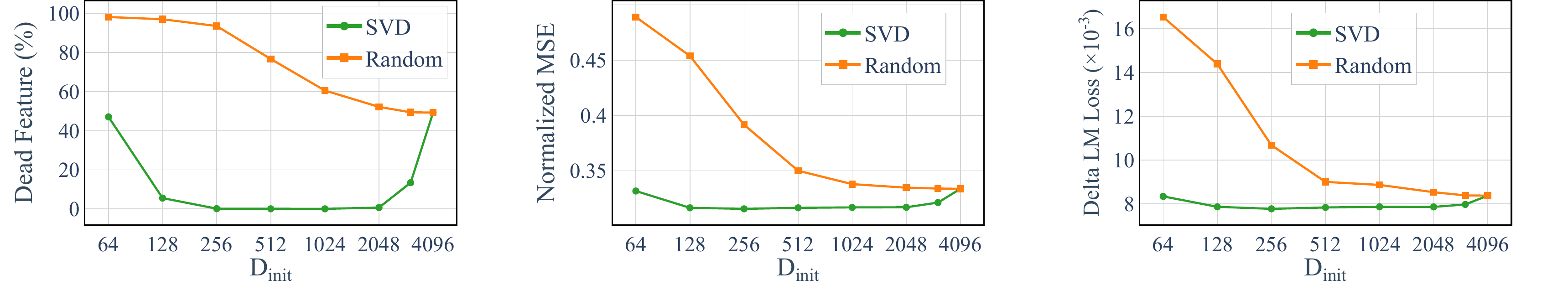}
    \caption{For activations with a full space dimension of 4096, proportion of dead features~(left), normalized MSE~(mid) and Delta LM loss~(right) across different subspace dimensions. Random subspaces are used as the baseline, whereas only initialization with the active subspace yields improvement.}
    \label{fig:active_subspace_init}
\end{figure}

We employ random subspace initialization as a baseline and observe that it consistently degrades SAE training across all metrics, as shown in Figure~\ref{fig:active_subspace_init}.

\subsection{Apply on Near-Full-Rank Activation}
\label{appendix:ablation_study:asi_on_resid}

\begin{figure}[htbp]
    \centering
    \includegraphics[width=\textwidth]{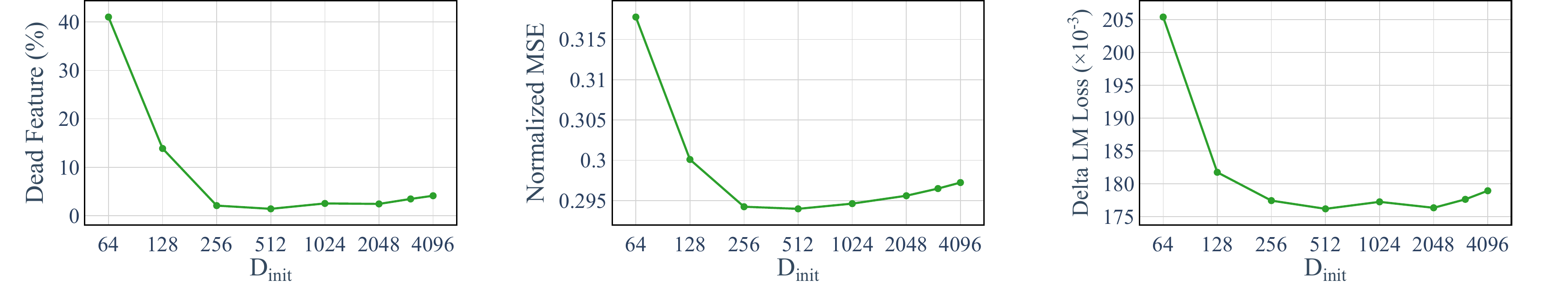}
    \caption{For activations with a full space dimension of 4096, proportion of dead features~(left), normalized MSE~(mid) and Delta LM loss~(right) across different subspace dimensions. Random subspaces are used as the baseline, whereas only initialization with the active subspace yields improvement.}
    \label{fig:d_init_resid}
\end{figure}

We also apply Active Subspace Initialization (ASI) to near-full-rank activations, such as those in the residual stream, to evaluate its generality. When training an SAE on the post-layer-15 residual stream of Llama-3.1-8B, we find ASI yields minimal gains~(Figure~\ref{fig:d_init_resid}). This is consistent with our expectation, as these activations inherently exhibit a lower rate of dead features even with standard initialization.

\subsection{Choice of the Initial Dictionary Size $d_{init}$}
\label{appendix:ablation_study:d_init}

As shown in Figure~\ref{fig:d_init}, $d_{init}$ is a hyperparameter with a wide range of acceptable values (from 256 to 2048). We hypothesize that the performance degradation when $d_{init}$ is very low is due to a combination of the dictionary failing to cover the subspace containing the key information needed for effective reconstruction and the features being too crowded.

\section{Pseudo-code for implementing Active Subspace Init}
\label{appendix:pseudo_code}

Below is a PyTorch-style pseudo-code for Active Subspace Initialization.

\paragraph{Use on SAE}

\begin{verbatim}
# X: activation batch [batch_size, d_model]
# W_E: decoder weight [d_model, d_sae]
# W_D: decoder weight [d_sae, d_model], initialized uniformly
# d_active_subspace: target subspace dimension

# 1. Demean the activations
demeaned_X = X - X.mean(dim=0) # [batch_size, d_model]

# 2. Compute SVD
U, S, V = torch.svd(demeaned_label) # V: [d_model, d_model]

# 3. Take top-d_init singular vectors
proj_weight = V[:, :d_init]  # [d_model, d_init]

# 4. Fold projection into decoder weights
W_D.copy_(W_D @ proj_weight @ proj_weight.T)

# 5. Init W_E with W_D.T
W_E.copy_(W_D.T)
\end{verbatim}

\paragraph{Use on Lorsa}

\begin{verbatim}
# Input:
# X: input activations [b, s, d]  (b=batch_size, s=seq_len, d=d_model)
# mhsa: pretrained MHSA module
#   mhsa.W_V: [n, d, h]  (n=n_heads, h=d_head, note: n*h=d)
#   mhsa.W_O: [n, h, d]
# Lorsa parameters to initialize:
#   W_V: [n_lorsa, d]  (n_lorsa = number of Lorsa heads)
#   W_O: [n_lorsa, d]
# d_qk: Lorsa head dimension for initialization

# 1. Compute per-head V projections
X_flat = X.reshape(b*s, d)                         # [b*s, d]
W_V_cat = mhsa.W_V.permute(1,0,2).reshape(d, d)    # [d, d]
V_per_head = (X_flat @ W_V_cat).reshape(b*s, n, h) # [b*s, n, h]

# 2. Project V back to d_model space for each head
# captured_v[:, i, :] = V_per_head[:, i, :] @ mhsa.W_V[i].T
captured_v = einsum('bnh, nhd -> bnd', 
                    V_per_head, mhsa.W_V.permute(0,2,1))
# captured_v: [b*s, n, d]

# 3. Initialize Lorsa heads from each original head's active subspace
rate = n_lorsa // n
for i in range(n):
    slice_i = [rate*i : rate*(i+1)]
    
    # 3.1 Extract this head's captured V
    v = captured_v[:, i, :]            # [b*s, d]
    
    # 3.2 Demean
    demeaned_v = v - v.mean(dim=0)     # [b*s, d]
    
    # 3.3 SVD on transposed data to get principal directions
    U, S, _ = svd(demeaned_v.T)        # demeaned_v.T: [d, b*s]
                                       # U: [d, d]
    
    # 3.4 Take top-d_qk principal directions as projection
    proj = U[:, :d_qk]                 # [d, d_qk]
    
    # 3.5 Update W_V: project from initial d_qk space to principal subspace
    W_V[slice_i] = W_V[slice_i, :d_qk] @ proj.T    # [rate, d]
    
    # 3.6 Update W_O: chain updated W_V through original head's OV circuit
    # OV_i = mhsa.W_V[i] @ mhsa.W_O[i]: [d, h] @ [h, d] = [d, d]
    W_O[slice_i] = W_V[slice_i] @ mhsa.W_V[i] @ mhsa.W_O[i]
    # [rate, d] @ [d, h] @ [h, d] = [rate, d]

# 4. Normalize all Lorsa weights (row-wise)
W_V = W_V / W_V.norm(dim=1, keepdim=True)   # [n_lorsa, d]
W_O = W_O / W_O.norm(dim=1, keepdim=True)   # [n_lorsa, d]
\end{verbatim}

The strategy of initialize $W_O$ in Lorsa is a method like the \textbf{tied initialization} used in SAEs to ensure alignment between feature encoding and decoding\footnote{"Match" means encoder can be initialized to predict relatively accurate feature activation values for decoder.}. This approach has been shown to be crucial for reducing dead features in SAEs~\citep{gao2024oaisae}. We think the same thought could also be used to improve the replacement model for MLP~(trancoder and cross layer transcoder), which we leave a deeper investigation to future work.

\section{Use ASI on other Activation Functions}
\label{appeneix:asi_other_activation_fuction}

\subsection{Jumprelu}
\label{appeneix:asi_other_activation_fuction:jumprelu}

\begin{figure}[htbp]
    \centering
    \includegraphics[width=0.9\textwidth]{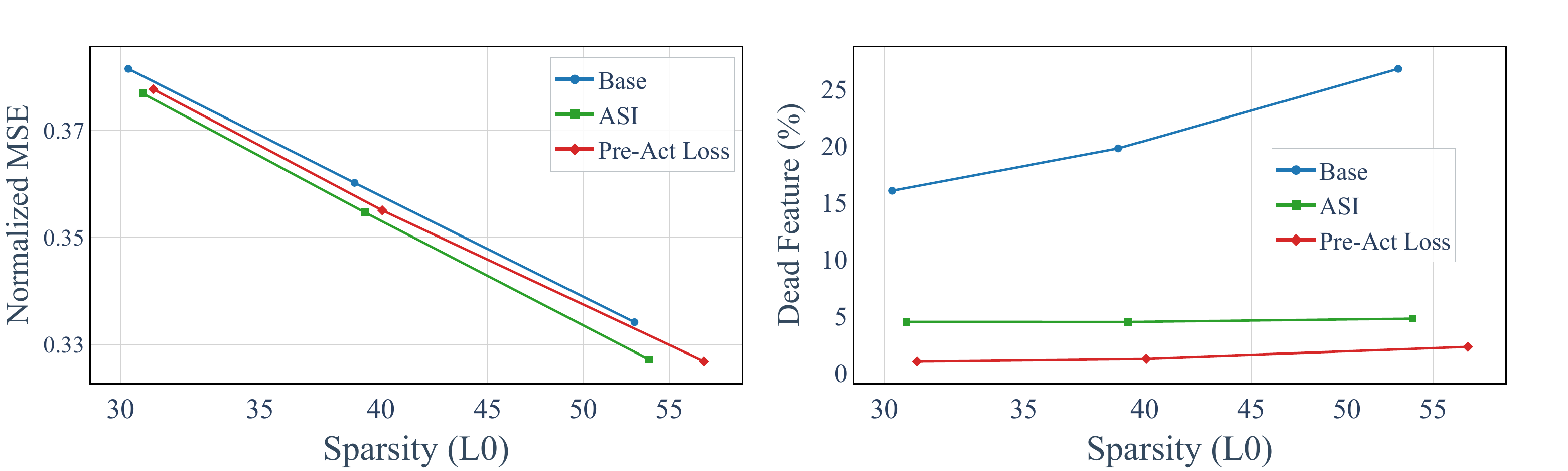}
    \caption{For the attention output of the middler layer of Llama-3.1-8B, using ASI on JumpRelu SAEs which has a low initial $\ell_0$ is effective. Details in Appendix~\ref{appeneix:asi_other_activation_fuction:jumprelu}}
    \label{fig:jumprelu_low_l0}
\end{figure}

\begin{figure}[htbp]
    \centering
    \includegraphics[width=0.9\textwidth]{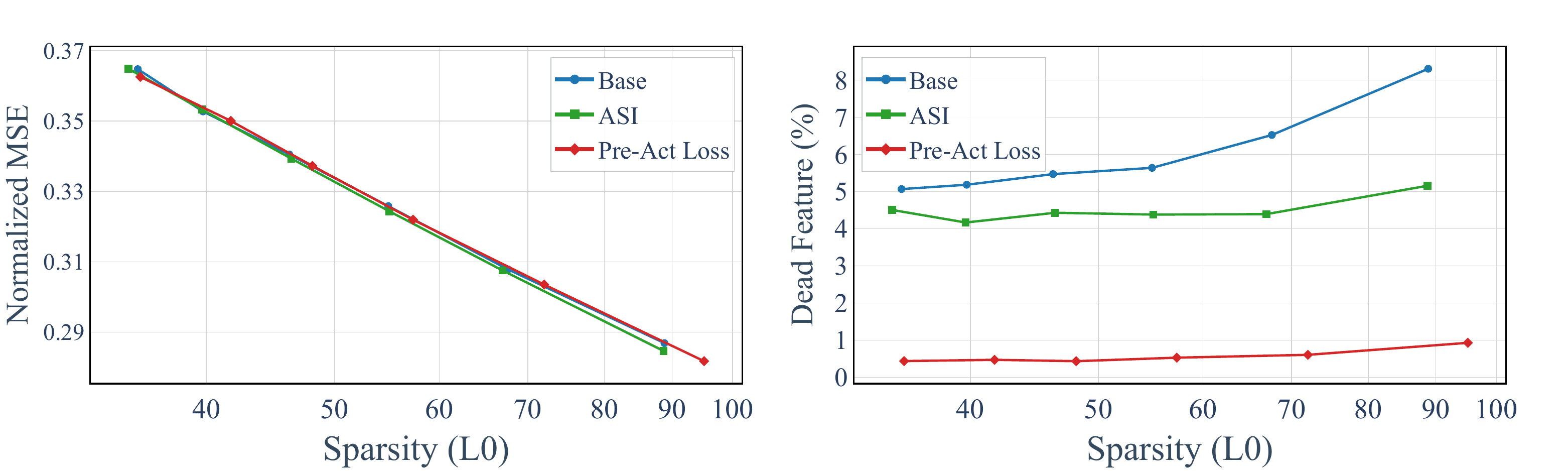}
    \caption{For the attention output of the middler layer of Llama-3.1-8B, using ASI on JumpRelu SAEs which has a high initial $\ell_0$ than gradually decreasing shows little improvement. Details in Appendix~\ref{appeneix:asi_other_activation_fuction:jumprelu}}
    \label{fig:jumprelu_high_l0}
\end{figure}

Another wildly used activation fuction is Jumprelu~\citep{rajamanoharan2024jumprelusae}. We trained the Jumprelu SAEs under two different hyperparameter settings: one with a consistently low $\ell_0$ value and another where $\ell_0$ gradually decreases from a higher initial value, as described in Appendix~\ref{appendix:sae_training_details:jumprelu_sae}. We observed that our method is effective in the former case~(Figure~\ref{fig:jumprelu_low_l0}) but shows little improvement in the latter~(Figure~\ref{fig:jumprelu_high_l0}). We train these SAEs following Appendix~\ref{appendix:sae_training_details}.

For cases where one follows a schedule that gradually reduces $\ell_0$ from a high initial value, we recommend first applying PCA to reduce the dimensionality of the data. The SAE can then be trained on the reduced representation until the $\ell_0$ level reaches the target range. Afterwards, the PCA projection matrix can be folded into the model parameters, and training can continue in the original space. This achieves a similar effect without the drawbacks of prolonged training in the high-$\ell_0$ regime.

\subsection{TopK with K Anneal}
\label{appeneix:asi_other_activation_fuction:k_anneal}

\begin{figure}[htbp]
    \centering
    \includegraphics[width=0.9\textwidth]{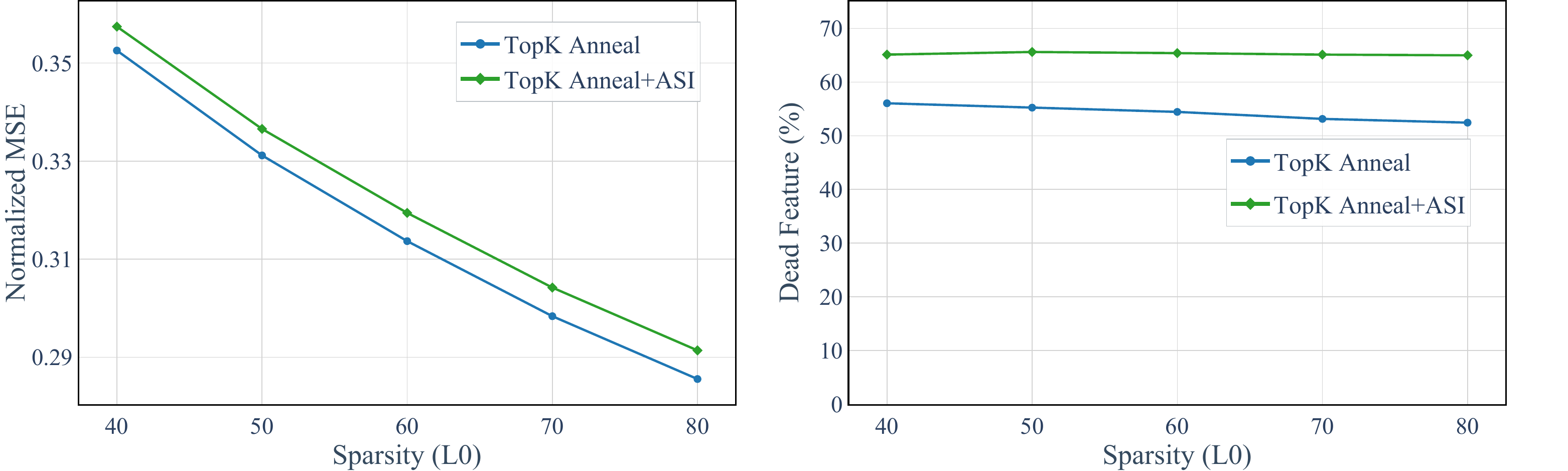}
    \caption{For the attention output of the middler layer of Llama-3.1-8B, using ASI on TopK SAEs which sets K to a high value and then lets it decrease during training fails. Details in Appendix~\ref{appeneix:asi_other_activation_fuction:k_anneal}}
    \label{fig:anneal_topk}
\end{figure}

To enhance the finding in Section~\ref{appeneix:asi_other_activation_fuction:jumprelu}, we conduct experiments on a variant of TopK, which sets K to a high value and then lets it decrease during training~\citep{he2024llamascope}. We find ASI also fails in this case~(Figure~\ref{fig:anneal_topk}).

\section{Stale Momentum as Another Root Cause of Dead Features}
\label{appendix:stale_momentum}

Recent work by~\citet{brichen2023stale} identifies \emph{stale momentum} as a key cause to dead feature formation. Specifically, when a feature remains inactive over training steps, its associated optimizer momentum continues to accumulate. If the feature activates, the stale momentum results in disproportionately large updates, destabilizing training and potentially suppressing that feature permanently.

To directly address this, we adopt \emph{SparseAdam}, an optimizer tailored for sparse activation settings, designed for more efficient use of compute and memory. SparseAdam updates both parameters and moments only when the corresponding feature is active. This could effectively prevent the harmful accumulation of stale momentum. Empirically, we observe that this change substantially reduces the rate of dead feature formation in large-scale SAE training. We believe that this is a core technique for scaling sparse dictionary methods, as stale momentum is a common problem for them.

\section{Compare Features of SAEs with and without ASI}
\label{appendix:compare_feature}

\subsection{Monosemanticity}

We conducted an additional analysis to assess the degree of monosemanticity exhibited by features learned by the base TopK SAE and the ASI-enhanced SAE. Following the rubric of~\citet{cunningham2024rubric}, we performed a blinded evaluation of 100 randomly sampled features and recorded the semantic consistency scores assigned to each feature.

For clarity, we reproduce below the scoring rubric used for evaluating activation consistency:

\begin{itemize}
    \item \textbf{5}: Clear pattern with no deviating examples
    \item \textbf{4}: Clear pattern with one or two deviating examples
    \item \textbf{3}: Clear overall pattern but quite a few examples not fitting that pattern
    \item \textbf{2}: Broad consistent theme but lacking structure
    \item \textbf{1}: No discernible pattern
\end{itemize}

To provide transparency, Tables~\ref{tab:asi_vs_topk_scores} summarize the distribution of scores for TopK and ASI.

Across both models, we found no statistically significant differences in score distributions at this scale of analysis. Variations between the two SAE variants were marginal and did not indicate systematic differences in feature quality. This result is consistent with our expectations, as our method does not modify the SAE architecture and is not designed to intervene in how features are formed. Consequently, the potential risk of degrading feature quality remains low.

\begin{table}[h]
\centering
\caption{Monosemanticity score distributions for ASI-enhanced SAE and base TopK SAE features.}
\label{tab:asi_vs_topk_scores}
\begin{subtable}[t]{0.48\textwidth}
    \centering
    \caption{ASI-enhanced SAE}
    \begin{tabular}{lc}
        \toprule
        \textbf{Score} & \textbf{Count} \\
        \midrule
        5 & 17 \\
        4 & 14 \\
        3 & 6 \\
        2 & 5 \\
        1 & 8 \\
        \bottomrule
    \end{tabular}
\end{subtable}
\hfill
\begin{subtable}[t]{0.48\textwidth}
    \centering
    \caption{Base TopK SAE}
    \begin{tabular}{lc}
        \toprule
        \textbf{Score} & \textbf{Count} \\
        \midrule
        5 & 18 \\
        4 & 12 \\
        3 & 6 \\
        2 & 6 \\
        1 & 8 \\
        \bottomrule
    \end{tabular}
\end{subtable}
\end{table}

\subsection{Analysis of SAE Features in the Dead Subspace}

To assess whether ASI alters the SAE's behavior in directions corresponding to the dead subspace, we perform a comparative analysis between Attention Output SAEs trained with and without ASI under identical configurations (same number of features, $K$, and all other hyperparameters).

\paragraph{Feature alignment with the dead subspace.}
We compute the cosine similarity between each SAE feature and the dead subspace, restricting the analysis to alive features (which accounts for the difference in total counts). The distribution of cosine values is summarized in Table~\ref{tab:dead-subspace-cosine}. Across all intervals, the ASI-initialized SAE exhibits a larger number of alive features, while both methods exhibit very few features that align closely with the dead subspace. This suggests two possible explanations: (i) features in the dead subspace have extremely small magnitude and provide insufficient signal for the SAE to learn, or (ii) the dead subspace does not contain meaningful standalone features, and only small components of features reside in this region.

\begin{table}[h]
\centering
\caption{Distribution of cosine similarity between SAE features and the dead subspace (alive features only).}
\label{tab:dead-subspace-cosine}
\begin{tabular}{c|cccc}
\toprule
Method & {[0.0, 0.05)} & {[0.05, 0.1)} & {[0.1, 0.15)} & {[0.15, 1]} \\
\midrule
TopK & 10176 & 189 & 6 & 0 \\
ASI  & 24576 & 936 & 8 & 0 \\
\bottomrule
\end{tabular}
\end{table}

\paragraph{Reconstruction error in the dead subspace.}
We further project the reconstruction error onto the dead subspace to quantify the SAE’s reconstruction performance on components lying in this region. As shown in Table~\ref{tab:dead-subspace-mse}, the reconstruction errors are nearly identical between the two methods, with the ASI showing only a marginal improvement.

\begin{table}[h]
\centering
\caption{Reconstruction error projected onto the dead subspace.}
\label{tab:dead-subspace-mse}
\begin{tabular}{c|c}
\toprule
Method & MSE in dead subspace \\
\midrule
TopK & 0.00350 \\
ASI  & 0.00334 \\
\bottomrule
\end{tabular}
\end{table}

Overall, these analyses indicate that ASI has minimal impact on the SAE's behavior within the dead subspace, while substantially reducing the number of dead features.

\section{Lorsa Implementation Details}
\label{appendix:lorsa_details}

\subsection{Hyperparameters}

\paragraph{Model, Dataset, Layer} Llama-3.1-8B, SlimPajama, 15(index start at 0).
\paragraph{Dictionary Size} We empirically set the num of Lorsa heads $n_{heads}^{Lorsa}=32768$ which is $8\times d_{model}$. We set the num of QK group of Lorsa $n_{qk}^{Lorsa}=256$ which is $8\times n_{heads}^{MHSA}$. We set the dimension of QK of Lorsa $d_{qk}^{Lorsa}=128$ which is $d_{qk}^{MHSA}$.
\paragraph{Batch Size} We empirically set the batch size to $32768$.
\paragraph{Optimizer} We use the Adam optimizer, with $\beta_1 = 0.9$, $\beta_2 = 0.999$, and $\epsilon = 10^{-8}$.
\paragraph{Learning Rate} The learning rate is sweeped in [$1\mathrm{e}{-5}$, $2\mathrm{e}{-5}$, $4\mathrm{e}{-5}$, $6\mathrm{e}{-5}$, $8\mathrm{e}{-5}$, $1\mathrm{e}{-4}$, $2\mathrm{e}{-4}$, $4\mathrm{e}{-4}$], and we ultimately use $2\mathrm{e}{-4}$. We employ a three-phase learning rate schedule consisting of a linear warm-up, a stable phase, and a linear decay. The learning rate increases linearly from zero to its maximum value over the first 500 steps, remains constant during the intermediate phase, and then decays linearly to 1\% of the maximum value over the final 20\% of the total training steps.

\paragraph{AuxK} We follow \citet{gao2024oaisae} to set auxiliary loss coefficient \(\alpha\) as $\frac{1}{32}$. We sweep the $k_{aux}$ in [256, 512, 1024, 2048] and finally choose 512. 

\paragraph{Dimension of Subspace for SAE Initialization~($d_{init}$)} Because the active subspace of the input and output of each MHSA heads is very close to the dimension of MHSA head~($d_{head}$), we set it directly to $d_{head}$. We found that increasing or decreasing this value did not improve performance. 

\paragraph{Total Tokens} We use $800$M tokens for each Lorsa training.

\paragraph{Sequence Length} We truncate each document to 2048 tokens. During training, we exclude the activations corresponding to
the \textless bos\textgreater~, \textless eos\textgreater and \textless pad\textgreater~tokens.

\subsection{Initialization}

We initialize the query and key matrices $W_Q$ and $W_K$ using Xavier uniform initialization \citep{glorot2010understanding}. The value matrix $W_V$ is initialized from a normal distribution $\mathcal{N}(0, 1/\sqrt{d_{sae}})$, while the output matrix $W_O$ is initialized from $\mathcal{N}(0, 1/\sqrt{d_{model}})$. All bias terms $b_Q$, $b_K$, $b_V$, and $b_D$ (if used) are initialized to zero.


\end{document}